\documentclass{article} 
\usepackage{iclr2027_conference,times}

\usepackage{amsmath,amsfonts,bm}

\def\eqref#1{equation~\ref{#1}}

\def\1{\bm{1}}

\DeclareMathAlphabet{\mathsfit}{\encodingdefault}{\sfdefault}{m}{sl}
\SetMathAlphabet{\mathsfit}{bold}{\encodingdefault}{\sfdefault}{bx}{n}

\usepackage{graphicx}
\usepackage{booktabs}
\usepackage{array}
\usepackage{multirow}
\usepackage{tikz}
\usepackage{colortbl}
\usepackage{hyperref}
\usepackage{placeins}
\usepackage{url}

\title{MetaKernelBench: Measuring GPU Kernel Knowledge Transfer Beyond Code}

\author{%
\textbf{Xueyi Chen\textsuperscript{1}\thanks{Equal contribution.\quad$^{\dagger}$Corresponding author.} \quad Shiyu Liu\textsuperscript{1,2}\footnotemark[1] \quad Xin Jin\textsuperscript{1} \quad Yuhua Zheng\textsuperscript{1,3} \quad Xin Li\textsuperscript{1,3}} \\
\textbf{Haolei Bai\textsuperscript{1} \quad Junhan Zhu\textsuperscript{1} \quad Huan Wang\textsuperscript{1}\rlap{$^{\dagger}$}} \\
\textsuperscript{1}Westlake University \quad \textsuperscript{2}Shanghai Jiao Tong University \quad \textsuperscript{3}Zhejiang University \\
\texttt{\{chenxueyi,wanghuan\}@westlake.edu.cn} \\
\url{https://yige24.github.io/MetaKernelBench}
}

\iclrfinalcopy 
\begin{document}

\vspace*{-36pt}
\maketitle
\lhead{Preprint.}
\enlargethispage{11pt}
\vspace{-18pt}

\begin{abstract}
Recent GPU kernel optimization agents retain what they learn in knowledge bases or as distilled skills.
Kernel benchmarks score each attempt's implementation for correctness and speed but leave the reuse value of retained experience unmeasured.
We introduce MetaKernelBench, which measures whether experience distilled from an attempt in one kernel domain-specific language (DSL) improves a fresh attempt at the same problem in another.
Its 74 problems are fused subgraphs in six families, each posed as a pair of CuTe DSL and TIRx variants that differ only in the DSL.
The agent first attempts each variant solo and is instructed to distill what it learns into a natural-language skill, which is transferred whether or not the source attempt passes verification.
The skill is the only extra input to a skill-conditioned attempt by the same model in the other DSL.
We compare each skill-conditioned attempt with the solo attempt on the same variant under matched per-attempt budgets, scoring correctness and end-to-end runtime.
Across six models and both directions, paired lift over solo attempts ranges from $-19$\% to $+29$\%.
Four models gain in both directions, yet regressions occur on 16\% to 45\% of problems in every model and direction.
Outcomes follow the source attempt's result relative to the target's solo attempt rather than source success alone, improving in 71\% of comparisons when the source stands above and regressing in 54\% when it stands below.
MetaKernelBench complements implementation-quality metrics by measuring same-problem cross-DSL kernel knowledge transfer.
\end{abstract}

\section{Introduction}
\label{sec:introduction}

GPU kernel optimization agents iteratively generate, profile, and revise implementations using execution feedback~\citep{wang2025geak,wei2025astra}.
Recent agents retain what they learn in persistent knowledge bases or distill successful rollouts into skills~\citep{dong2026kernelblaster,fu2026davincikernel}.
Kernel benchmarks score the implementation each attempt produces for correctness and speed relative to a reference~\citep{ouyang2025kernelbench}.
Newer benchmarks extend this scoring to other hardware platforms and kernel domain-specific languages (DSLs) and to unseen input configurations~\citep{wen2026multikernelbench,younesian2026agentkernelarena}.
Such per-attempt scores leave the reuse value of retained experience unmeasured.
We ask whether experience distilled from an attempt in one DSL can improve a fresh attempt at the same problem in another.

We use \emph{kernel knowledge} to denote problem-, hardware-, and strategy-level principles that guide implementation decisions and can be expressed independently of a DSL's syntax and APIs.
Examples include avoiding the materialization of intermediate tensors, overlapping data movement with computation, and dividing work between eager PyTorch and DSL kernels.
A \emph{skill} records an agent's findings in natural language, without guaranteeing that they are correct or useful to a later attempt.

Pairing the same problem across DSLs preserves the computational and hardware conditions that make prior experience relevant while changing how the kernel must be expressed.
Repeating the problem in the same DSL would let DSL-specific recipes carry over unchanged, whereas switching problems would change which findings remain relevant.
Advice about eager PyTorch may still apply regardless of the DSL, but kernel strategies must be re-implemented in the other DSL.
This setting does not test generalization to new problems, workloads, or hardware.

We introduce \textbf{MetaKernelBench}, which pairs CuTe DSL and TIRx variants of 74 fused subgraphs under fixed semantics, workloads, hardware, and verification rules.
During each variant's solo attempt without transferred knowledge, the agent is instructed to distill optimization findings into a skill.
Every available source skill is transferred whether or not the source attempt passes verification, so failed attempts can also contribute useful findings.
Each skill is the sole additional input to a fresh skill-conditioned attempt by the same model in the other DSL, with the source implementation and trajectory withheld.
We compare each skill-conditioned attempt with the solo attempt on the same variant under matched agent configurations and per-attempt budgets.
Paired comparisons assess the distillation and use of skills jointly by measuring the correctness and end-to-end runtime of implementations that may combine eager PyTorch and DSL kernels.

In single-attempt comparisons across six models, paired lift ranges from $-19$\% to $+29$\% by model and direction.
Four models gain in both directions, yet every model and direction has regressions on 16\% to 45\% of problems.
Outcomes follow the source attempt's pass and speed tier relative to the target's solo attempt rather than source success alone.
When the source stands above the target's solo attempt, 71\% of comparisons improve, and when it stands below, 54\% regress.
Source skills with unsupported conclusions that rule out alternatives are more common in harmed than in helped comparisons for five of six models, while other content associated with improvements varies across models.
A failure analysis finds that larger models are mostly led astray by skills they follow, whereas the two models with the lowest pooled solo pass rates mostly fail in their own execution.

Our contributions are as follows.
\begin{itemize}
    \item We formulate same-problem cross-DSL kernel knowledge transfer as an evaluation dimension complementary to the correctness and performance of an individual implementation.
    \item We introduce MetaKernelBench, which pairs 74 problems across CuTe DSL and TIRx, transfers only the source skill regardless of source success, and compares skill-conditioned attempts with solo attempts under matched per-attempt budgets.
    \item We characterize transfer across six models and both directions, show that outcomes follow the source attempt's standing relative to the target's solo attempt, and relate them to recorded guidance through content, case, and failure analyses.
\end{itemize}

\begin{figure}[t]
    \centering
    \includegraphics[width=\textwidth]{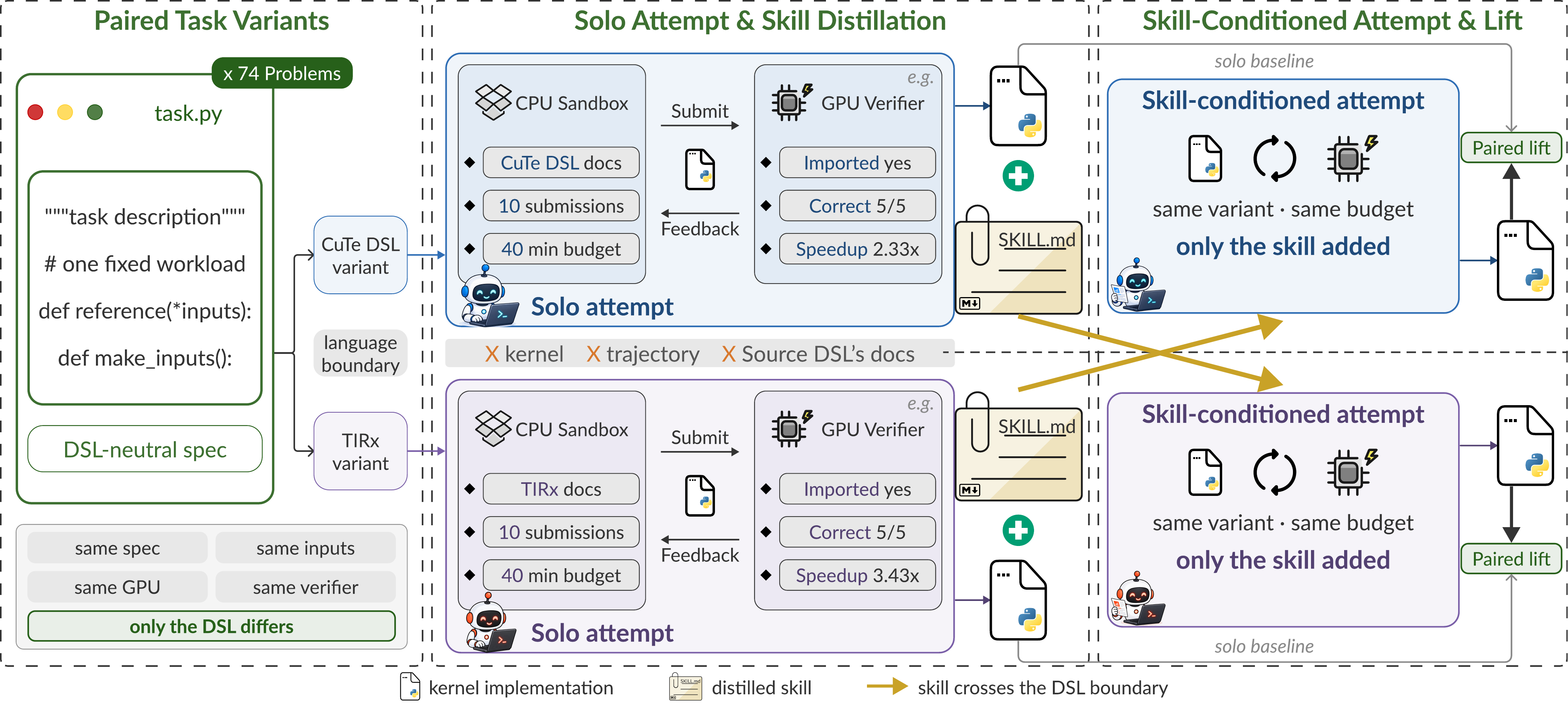}
    \caption{Overview of the MetaKernelBench pipeline.
        \textbf{Left:} one DSL-neutral specification yields paired CuTe DSL and TIRx variants.
        \textbf{Middle:} solo attempts use verifier feedback to optimize implementations and distill findings into natural-language skills.
        \textbf{Right:} only the source skill crosses to a fresh attempt in the other DSL, and paired lift compares its implementation with the solo attempt's.
    }
    \label{fig:pipeline}
\end{figure}

\section{MetaKernelBench}
\label{sec:metakernelbench}

MetaKernelBench pairs CuTe DSL and TIRx variants under a shared specification and grading contract, so paired variants differ only in the DSL.
It compares each skill-conditioned attempt with the solo attempt on the same variant under matched budgets, so the two differ only in the skill (Figure~\ref{fig:pipeline}).

\subsection{Task Formulation}
\label{sec:task_formulation}

Each problem $p$ has a DSL-neutral executable specification containing a task description, a PyTorch reference function, and a generator of fresh inputs for a fixed workload configuration.
A shared grading contract checks correctness against the reference and measures end-to-end speedup over the reference's eager PyTorch execution.
For each kernel DSL $d\in\mathcal{D}=\{\mathrm{CuTe},\mathrm{TIRx}\}$, the task variant $\tau_{p,d}$ adds only $d$ and its software environment to this specification.
The objective is to find the fastest valid implementation using $d$ on a single NVIDIA B200 GPU.

A valid implementation may combine eager PyTorch, ordinary Python, and kernels written in $d$.
Every invocation must execute a DSL kernel that performs a required computation or data-movement step on which at least one returned output tensor depends.
A DSL step fails the \emph{bypass test} if bypassing it and returning the existing eager result still yields valid outputs.
Other kernel languages and compilers, CUDA Graphs, calls to the reference, and kernels imported from DSL kernel libraries are forbidden.
Both DSLs are embedded in Python, so hybrid implementations are their usual form, and the boundary between eager and DSL work is itself an optimization decision.
End-to-end timing covers each complete invocation, including Python dispatch, eager operations, and DSL kernels.

CuTe DSL and TIRx both give kernel authors explicit control over pipeline structure, memory placement, synchronization, and hardware roles.
CuTe DSL uses layout algebra to partition data and computation across threads~\citep{cutlass}, whereas TIRx uses storage layouts to guide compiler dispatch of tile primitives~\citep{tvm2026tirx}.
Kernel code written in one DSL must be re-expressed in the other, while eager PyTorch code runs unchanged in both variants.

\subsection{Transfer Protocol}
\label{sec:transfer_protocol}

A shared directive instructs the agent to maintain a natural-language skill throughout each attempt.
The skill should record evidence-backed optimization findings, work partitioning between eager PyTorch and DSL kernels, reasons for failures, applicability conditions, and open questions.
The agent is told that this skill may guide a fresh attempt on the same problem in another DSL.
Conclusions should not depend on source-DSL syntax or APIs, although details needed to explain evidence are allowed.
The directive forbids code, commands, and API instructions.

For source and target DSLs $d_s\neq d_t$ in $\mathcal{D}$, the attempt outputs for direction $d_s\!\rightarrow d_t$ are:
\begin{equation}
(k^{\mathrm{solo}}_{p,d},\,\sigma_{p,d})\leftarrow \mathcal{A}(\tau_{p,d},\,\emptyset),\quad d\in\mathcal{D},
\qquad
(k^{\mathrm{skill}}_{p,\,d_s\rightarrow d_t},\,\cdot\,)\leftarrow \mathcal{A}(\tau_{p,d_t},\,\sigma_{p,d_s}),\quad \sigma_{p,d_s}\neq\emptyset.
\label{eq:transfer_arms}
\end{equation}
In Equation~\ref{eq:transfer_arms}, $\leftarrow$ denotes a fresh execution of agent procedure $\mathcal{A}$ with a fixed model, agent configuration, and per-attempt budget.
$k$ and $\sigma$ denote the final implementation and skill, respectively, $\emptyset$ denotes absence, and the dot stands for the target's own skill, which is not transferred.
Each solo attempt supplies the source skill in one direction and is the solo comparator in the other.

The target variant's solo and skill-conditioned attempts are the \emph{baseline} and \emph{treatment}, respectively.
The source skill is appended verbatim to the baseline instruction under a neutral header attributing it to a prior attempt on the same problem in another DSL.
Every available source skill is transferred whether or not the source implementation passes verification.
Transfer does not filter skill content, so the requested format is not enforced.
If the source skill is missing or blank, the treatment is not run but is recorded as skipped and inherits the baseline outcome for evaluation.

The baseline and treatment share the same task variant, verifier, feedback mechanism, and implementation rules.
The skill attachment is the only additional task input, with the source implementation and trajectory withheld.
Matching target budgets does not equalize total computation, since producing the transferred skill also requires a separate source attempt with the same 40-minute budget.
Eager PyTorch advice may apply directly, work partitioning may require revision, and kernel strategies require new target-DSL implementations.
Paired lift measures the combined end-to-end outcome without isolating the contribution of each type of guidance.

\begin{figure}[t]
    \centering
    \includegraphics[width=\textwidth]{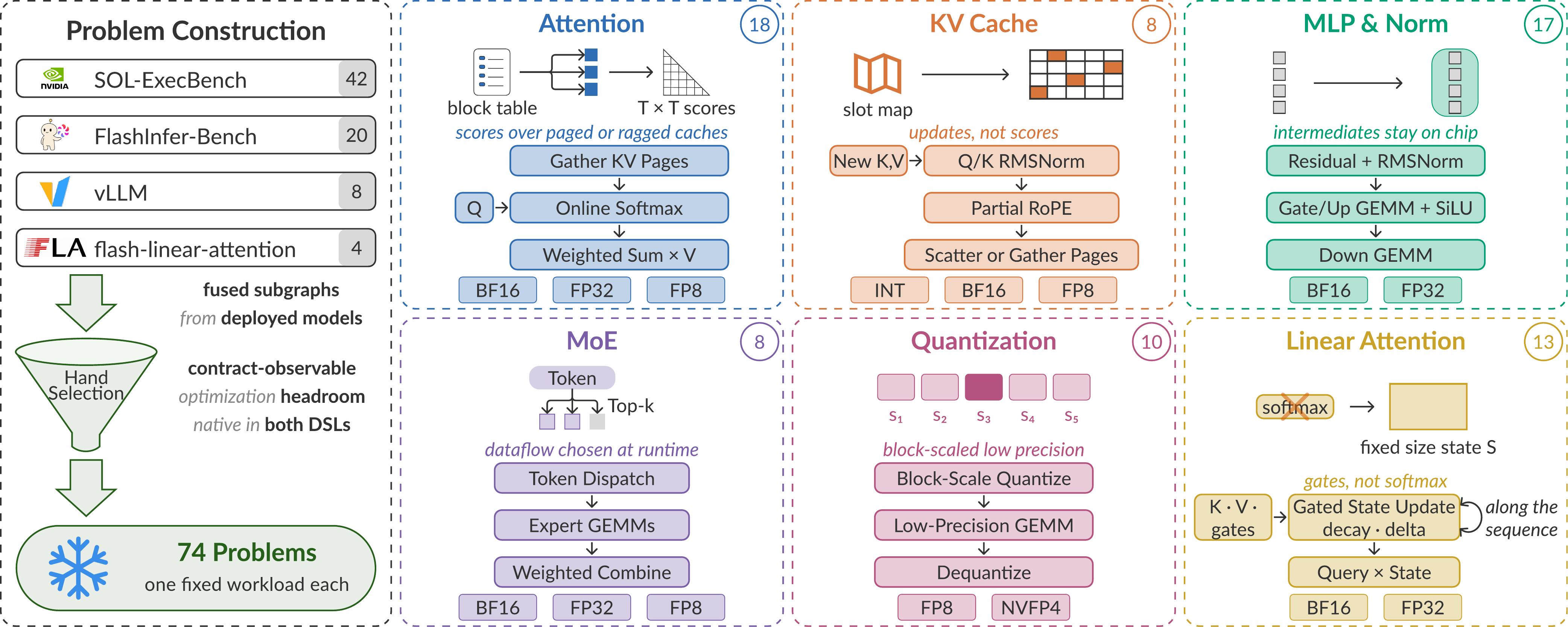}
    \caption{Construction and composition of the MetaKernelBench problem set.
        \textbf{Left:} manual screening of fused subgraphs from four upstream sources yields 74 fixed-workload problems.
        \textbf{Right:} six families show problem counts, representative computation patterns, and numerical formats.
    }
    \label{fig:dataset}
\end{figure}

\subsection{Problem Construction}
\label{sec:problem_construction}

MetaKernelBench uses fused subgraphs traceable to production models, serving stacks, or published reference implementations.
These subgraphs expose choices about fusion, intermediate materialization, and the division of work between eager PyTorch and DSL kernels.
DSL optimization may involve shared-memory and register use as well as overlapping data movement with computation.

A candidate is retained only if its defining behavior remains observable under the grading contract, it offers meaningful kernel-optimization headroom, and native implementation is feasible in both DSLs.
Although subgraphs may span multiple operator types, each is assigned to one of six families according to its main implementation constraint (Figure~\ref{fig:dataset}).
Attention involves attention scoring, often with paged, ragged, or sparse layouts, whereas KV Cache covers cache maintenance without scoring.
MLP \& Norm covers normalization and dense fusion chains, while MoE covers routing, dispatch, and expert computation.
Quantization centers on block-scaled low-precision computation, while Linear Attention covers recurrent, chunk-scanned, or convolutional sequence mixing.

The 74-problem catalog includes 42 problems adapted from SOL-ExecBench~\citep{lin2026solexecbench} and 20 from FlashInfer-Trace and FlashInfer-Bench~\citep{xing2026flashinferbench}.
The catalog contains 8 cache-maintenance problems constructed from vLLM tests and kernels~\citep{kwon2023vllm} and 4 linear-attention problems authored from official model configurations and tests in flash-linear-attention~\citep{yang2024fla}.
Each problem fixes a workload configuration and input distribution, selecting the largest admissible workload when multiple upstream records exist.
Where needed, we adapt upstream definitions to expose core behavior, stabilize discrete selections, support reference evaluation with promoted floating-point inputs, and complete standalone call contracts.

\section{Experimental Setup}
\label{sec:experimental_setup}

\subsection{Models and Agent Scaffold}
\label{sec:models_and_agent_scaffold}

We evaluate six models, namely DeepSeek-V4-Pro~\citep{deepseek2026v4}, GLM-5.3~\citep{zai2026glm53}, GLM-5.3-Flash~\citep{zai2026glm53flash}, Gemini 3.7 Flash~\citep{google2026gemini37}, GPT-5.6~Sol~\citep{openai2026gpt56}, and Qwen3.8-Flash~\citep{qwen2026flash}.
The panel contains three Flash-tier models and three larger models, with GLM-5.3 and GLM-5.3-Flash forming a same-family pair.
All models use Terminal-Bench's terminal-only Terminus-2 scaffold~\citep{merrill2026terminalbench} through Harbor 0.20.0~\citep{harbor}.
Context compression reintroduces the complete task instruction, including any attached source skill.
We do not override sampling temperature or reasoning effort in either arm.

\subsection{Execution Environment}
\label{sec:execution_environment}

Each attempt runs in an isolated CPU-only sandbox with two cores, 8 GB of memory, and no network access.
The sandbox provides the task specification, grading contract, submission script, fixed-version dependencies for the variant's DSL, and offline documentation.
The agent has a 40-minute wall-clock budget, including feedback waits, and at most 10 feedback submissions.
Each submission snapshots the implementation for evaluation in a fresh container on an NVIDIA B200, using new input seeds and the sandbox's DSL dependencies.
Feedback contains import and correctness results, errors, and, on success, the end-to-end implementation and eager-reference times with their ratio.
At budget expiry or an early stop, any collected final implementation is evaluated afresh on a B200.
Each phase permits up to three tries for transient infrastructure failures, with execution retries restarting the agent and final-evaluation retries re-evaluating only the collected implementation.

\subsection{Evaluation Metrics}
\label{sec:evaluation_metrics}

A scored pass requires float64-anchored correctness and completed end-to-end timing.
The verifier compares floating-point outputs against a reference evaluated with promoted inputs, using the original reference's error to set tolerances.
Each evaluation uses five correctness draws that must all pass, 10 warmup calls, and 50 timed calls, with correctness rechecked on three timed outputs.
A static check rejects implementations using TIRx's inline CUDA source entry point, which embeds raw CUDA.
The size of the DSL contribution is not scored, since runtime already reflects its value.

Solo metrics use final outcomes by model and DSL, including attempts that produced no skill.
Speedup is the eager-reference median runtime divided by the implementation's median runtime, capped at $100\times$.
We report pass rate $P$, passing-only geometric mean speedup (GM), and the $2\times$-or-better pass rate $R(2)$, retaining failures in both rate denominators.

A comparison $i$ is resolved when its source and both target arms have final outcomes, with a skipped treatment inheriting the baseline outcome.
For arm $a\in\{\mathrm{base},\mathrm{treat}\}$, let $p_i^a\in\{0,1\}$ indicate passing and $t_i^a$ be its end-to-end median runtime when passing, giving the paired ratio:
\begin{equation}
    \rho_i=\operatorname{clip}\bigl(t^{\mathrm{base}}_i/t^{\mathrm{treat}}_i,\,\tfrac14,\,4\bigr)\ \text{if } p_i^{\mathrm{base}}p_i^{\mathrm{treat}}=1,
    \qquad
    \rho_i=4^{\,p_i^{\mathrm{treat}}-p_i^{\mathrm{base}}}\ \text{otherwise}.
    \label{eq:paired_ratio}
\end{equation}
Equation~\ref{eq:paired_ratio} uses raw runtimes, not capped speedups, for jointly passing arms and assigns $4$ to failure-to-pass rescues, $1/4$ to pass-to-failure breaks, and $1$ to joint failures or skipped treatments.
The transition values are scores rather than measured speedups, placed at the clip bound so that a verdict change weighs as much as the largest retained runtime change.

Over a set $G$ of resolved comparisons, we aggregate $\rho_i$ as paired lift, help rate, and harm rate:
\begin{equation}
    \Delta\mathrm{GM}_G=\exp\Bigl(\frac{1}{|G|}\sum_{i\in G}\ln\rho_i\Bigr)-1,
    \quad
    T_G=\frac{1}{|G|}\sum_{i\in G}\1[\rho_i\geq\tfrac54],
    \quad
    H_G=\frac{1}{|G|}\sum_{i\in G}\1[\rho_i\leq\tfrac45].
    \label{eq:speedup_lift}
\end{equation}
Equation~\ref{eq:speedup_lift} weights comparisons equally, leaving a neutral share of $1-T_G-H_G$.
We report $\Delta\mathrm{GM}_G$, $T_G$, and $H_G$ as percentages by model and direction, within each family and over all 74 problems.

\section{Results}
\label{sec:results}

\begin{figure}[t]
    \centering
    \includegraphics[width=\textwidth]{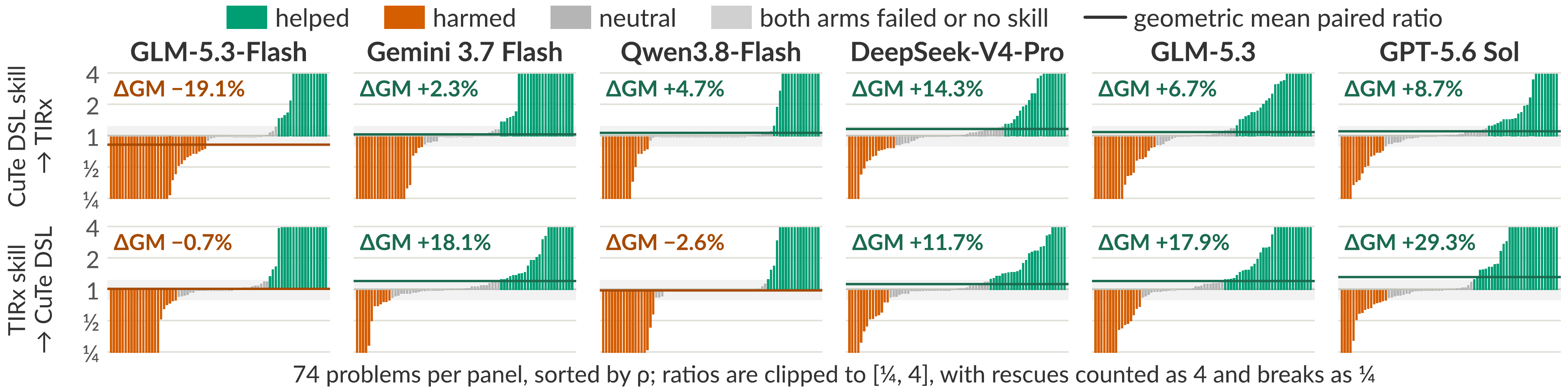}
    \caption{Problem-level paired ratios by model and transfer direction.}
    \label{fig:paired_lift_by_problem}
\end{figure}

\definecolor{liftpos}{RGB}{0,158,115}
\definecolor{liftneg}{RGB}{213,94,0}
\newcommand{\liftpad}{2mm}
\newcommand{\liftheight}{4mm}
\newcommand{\liftstrong}{45}
\newcommand{\liftweak}{4}

\newcommand{\liftsep}{\kern-0.1em$\scriptstyle/$\kern-0.1em}
\newcommand{\liftlabel}[3]{$#1$\liftsep$#2$\liftsep$#3$}
\newlength{\liftwidth}
\newlength{\liftmeasure}
\newcommand{\liftfit}[1]{\settowidth{\liftmeasure}{#1}\ifdim\liftmeasure>\liftwidth\setlength{\liftwidth}{\liftmeasure}\fi}

\newcommand{\liftcell}[6]{%
  \begin{tikzpicture}[baseline=(t.base)]
    \useasboundingbox (0mm,-0.5*\liftheight) rectangle (\liftwidth,0.5*\liftheight);
    \pgfmathsetlengthmacro{\lifthelp}{#5*\liftwidth}%
    \pgfmathsetlengthmacro{\liftharm}{#6*\liftwidth}%
    \ifdim \lifthelp>0pt
      \shade[left color=liftpos!\liftstrong, right color=liftpos!\liftweak]
        (0mm,-0.5*\liftheight) rectangle (\lifthelp,0.5*\liftheight);\fi
    \ifdim \liftharm>0pt
      \shade[left color=liftneg!\liftweak, right color=liftneg!\liftstrong]
        (\liftwidth-\liftharm,-0.5*\liftheight) rectangle (\liftwidth,0.5*\liftheight);\fi
    \node (t) at (0.5*\liftwidth,0mm) [font=\normalsize]%
      {\liftlabel{#2}{#1{#3}}{#4}};
  \end{tikzpicture}%
}
\newcommand{\lift}{\liftcell{}}
\newcommand{\liftbest}{\liftcell{\underline}}
\newcommand{\liftnone}{\makebox[\liftwidth]{--}}

\liftfit{\liftlabel{0.0}{{\scriptstyle-}44.6}{62.5}}
\liftfit{\liftlabel{11.8}{{\scriptstyle-}0.3}{17.6}}
\liftfit{\liftlabel{11.8}{{\scriptstyle-}0.9}{29.4}}
\liftfit{\liftlabel{11.8}{{\scriptstyle-}9.8}{11.8}}
\liftfit{\liftlabel{12.5}{{\scriptstyle+}3.0}{12.5}}
\liftfit{\liftlabel{12.5}{{\scriptstyle+}4.1}{12.5}}
\liftfit{\liftlabel{12.5}{{\scriptstyle-}0.9}{25.0}}
\liftfit{\liftlabel{12.5}{{\scriptstyle-}10.8}{25.0}}
\liftfit{\liftlabel{12.5}{{\scriptstyle-}17.7}{25.0}}
\liftfit{\liftlabel{12.5}{{\scriptstyle-}40.9}{62.5}}
\liftfit{\liftlabel{15.4}{{\scriptstyle+}6.1}{15.4}}
\liftfit{\liftlabel{15.4}{{\scriptstyle-}19.3}{30.8}}
\liftfit{\liftlabel{16.7}{{\scriptstyle-}7.4}{22.2}}
\liftfit{\liftlabel{17.6}{{\scriptstyle-}13.6}{29.4}}
\liftfit{\liftlabel{17.6}{{\scriptstyle-}16.2}{35.3}}
\liftfit{\liftlabel{20.0}{{\scriptstyle+}13.4}{20.0}}
\liftfit{\liftlabel{20.0}{{\scriptstyle+}3.7}{20.0}}
\liftfit{\liftlabel{20.0}{{\scriptstyle+}4.0}{20.0}}
\liftfit{\liftlabel{20.0}{{\scriptstyle-}19.1}{30.0}}
\liftfit{\liftlabel{20.0}{{\scriptstyle-}25.6}{50.0}}
\liftfit{\liftlabel{21.6}{{\scriptstyle+}4.7}{21.6}}
\liftfit{\liftlabel{22.2}{{\scriptstyle+}6.3}{27.8}}
\liftfit{\liftlabel{22.2}{{\scriptstyle-}17.3}{50.0}}
\liftfit{\liftlabel{23.0}{{\scriptstyle-}19.1}{44.6}}
\liftfit{\liftlabel{23.1}{{\scriptstyle+}1.0}{23.1}}
\liftfit{\liftlabel{23.1}{{\scriptstyle-}19.4}{38.5}}
\liftfit{\liftlabel{23.1}{{\scriptstyle-}28.8}{53.8}}
\liftfit{\liftlabel{23.1}{{\scriptstyle-}5.7}{46.2}}
\liftfit{\liftlabel{23.1}{{\scriptstyle-}7.4}{30.8}}
\liftfit{\liftlabel{23.5}{{\scriptstyle+}0.3}{23.5}}
\liftfit{\liftlabel{24.3}{{\scriptstyle-}2.6}{24.3}}
\liftfit{\liftlabel{25.0}{{\scriptstyle+}2.2}{37.5}}
\liftfit{\liftlabel{25.0}{{\scriptstyle+}20.6}{25.0}}
\liftfit{\liftlabel{25.0}{{\scriptstyle+}4.9}{37.5}}
\liftfit{\liftlabel{25.0}{{\scriptstyle-}21.4}{37.5}}
\liftfit{\liftlabel{25.0}{{\scriptstyle-}28.4}{50.0}}
\liftfit{\liftlabel{25.0}{{\scriptstyle-}30.3}{50.0}}
\liftfit{\liftlabel{25.0}{{\scriptstyle-}37.8}{62.5}}
\liftfit{\liftlabel{25.0}{{\scriptstyle-}5.7}{25.0}}
\liftfit{\liftlabel{27.0}{{\scriptstyle-}0.7}{31.1}}
\liftfit{\liftlabel{28.4}{{\scriptstyle+}14.3}{21.6}}
\liftfit{\liftlabel{29.4}{{\scriptstyle+}20.6}{5.9}}
\liftfit{\liftlabel{30.0}{{\scriptstyle-}6.5}{30.0}}
\liftfit{\liftlabel{32.4}{{\scriptstyle+}8.7}{20.3}}
\liftfit{\liftlabel{33.3}{{\scriptstyle+}5.2}{16.7}}
\liftfit{\liftlabel{33.8}{{\scriptstyle+}18.1}{16.2}}
\liftfit{\liftlabel{33.8}{{\scriptstyle+}2.3}{31.1}}
\liftfit{\liftlabel{35.1}{{\scriptstyle+}11.7}{18.9}}
\liftfit{\liftlabel{35.1}{{\scriptstyle+}6.7}{28.4}}
\liftfit{\liftlabel{37.5}{{\scriptstyle+}14.0}{37.5}}
\liftfit{\liftlabel{37.5}{{\scriptstyle+}20.2}{0.0}}
\liftfit{\liftlabel{37.5}{{\scriptstyle+}27.9}{0.0}}
\liftfit{\liftlabel{37.5}{{\scriptstyle+}68.2}{0.0}}
\liftfit{\liftlabel{38.5}{{\scriptstyle+}32.7}{23.1}}
\liftfit{\liftlabel{38.5}{{\scriptstyle+}40.0}{15.4}}
\liftfit{\liftlabel{38.9}{{\scriptstyle+}20.2}{22.2}}
\liftfit{\liftlabel{38.9}{{\scriptstyle+}4.2}{33.3}}
\liftfit{\liftlabel{38.9}{{\scriptstyle+}7.5}{33.3}}
\liftfit{\liftlabel{39.2}{{\scriptstyle+}29.3}{21.6}}
\liftfit{\liftlabel{40.0}{{\scriptstyle+}43.0}{10.0}}
\liftfit{\liftlabel{40.0}{{\scriptstyle-}19.9}{40.0}}
\liftfit{\liftlabel{40.5}{{\scriptstyle+}17.9}{21.6}}
\liftfit{\liftlabel{41.2}{{\scriptstyle+}12.0}{5.9}}
\liftfit{\liftlabel{41.2}{{\scriptstyle+}18.3}{11.8}}
\liftfit{\liftlabel{44.4}{{\scriptstyle+}22.2}{11.1}}
\liftfit{\liftlabel{44.4}{{\scriptstyle+}26.0}{27.8}}
\liftfit{\liftlabel{44.4}{{\scriptstyle+}8.2}{33.3}}
\liftfit{\liftlabel{46.2}{{\scriptstyle-}3.2}{38.5}}
\liftfit{\liftlabel{47.1}{{\scriptstyle+}44.6}{11.8}}
\liftfit{\liftlabel{5.9}{{\scriptstyle-}1.8}{17.6}}
\liftfit{\liftlabel{5.9}{{\scriptstyle-}9.3}{29.4}}
\liftfit{\liftlabel{50.0}{{\scriptstyle+}11.8}{37.5}}
\liftfit{\liftlabel{50.0}{{\scriptstyle+}29.4}{20.0}}
\liftfit{\liftlabel{50.0}{{\scriptstyle+}32.2}{12.5}}
\liftfit{\liftlabel{50.0}{{\scriptstyle+}40.4}{33.3}}
\liftfit{\liftlabel{50.0}{{\scriptstyle+}84.4}{12.5}}
\liftfit{\liftlabel{53.8}{{\scriptstyle+}56.0}{0.0}}
\liftfit{\liftlabel{55.6}{{\scriptstyle+}64.8}{11.1}}
\liftfit{\liftlabel{60.0}{{\scriptstyle+}41.1}{20.0}}
\liftfit{\liftlabel{60.0}{{\scriptstyle+}43.4}{20.0}}
\liftfit{\liftlabel{61.5}{{\scriptstyle+}73.3}{15.4}}
\liftfit{\liftlabel{62.5}{{\scriptstyle+}59.3}{12.5}}
\liftfit{\liftlabel{62.5}{{\scriptstyle+}67.9}{25.0}}
\liftfit{\liftlabel{70.0}{{\scriptstyle+}99.1}{0.0}}
\addtolength{\liftwidth}{\liftpad}

\begin{table}[t]
\centering
\caption{Cross-DSL skill transfer against the target variant's solo attempts.
Cells report help rate $T$, paired lift $\Delta$GM, and harm rate $H$ in percent, with green bars showing $T$, orange bars showing $H$, and the gap the neutral share.
Underlining marks the highest $\Delta$GM in each direction and column.
}
\label{tab:cross_dsl_transfer}
\setlength{\tabcolsep}{1pt}
\renewcommand{\arraystretch}{1.1}
\resizebox{\linewidth}{!}{%
\begin{tabular}{l cccccc !{\hspace{\tabcolsep}\vrule\hspace{\tabcolsep}} c}
\toprule
\textbf{Model} & \textbf{Attention} & \textbf{KV Cache} & \textbf{Linear Attn} & \textbf{MLP \& Norm} & \textbf{MoE} & \textbf{Quant} & \textbf{All} \\
{\footnotesize $T$\liftsep$\Delta$GM\liftsep$H$} & {\footnotesize(18)} & {\footnotesize(8)} & {\footnotesize(13)} & {\footnotesize(17)} & {\footnotesize(8)} & {\footnotesize(10)} & {\footnotesize(74)} \\
\midrule
\multicolumn{8}{c}{\textit{CuTe DSL skill $\longrightarrow$ fresh TIRx attempt}} \\
\cmidrule(lr){1-8}
GLM-5.3-Flash & \lift{22.2}{{\scriptstyle-}17.3}{50.0}{0.222}{0.500} & \lift{25.0}{{\scriptstyle-}37.8}{62.5}{0.250}{0.625} & \lift{23.1}{{\scriptstyle-}28.8}{53.8}{0.231}{0.538} & \lift{17.6}{{\scriptstyle-}16.2}{35.3}{0.176}{0.353} & \lift{37.5}{{\scriptstyle+}14.0}{37.5}{0.375}{0.375} & \lift{20.0}{{\scriptstyle-}19.1}{30.0}{0.200}{0.300} & \lift{23.0}{{\scriptstyle-}19.1}{44.6}{0.230}{0.446} \\
Gemini 3.7 Flash & \liftbest{44.4}{{\scriptstyle+}26.0}{27.8}{0.444}{0.278} & \lift{12.5}{{\scriptstyle-}0.9}{25.0}{0.125}{0.250} & \lift{46.2}{{\scriptstyle-}3.2}{38.5}{0.462}{0.385} & \liftbest{23.5}{{\scriptstyle+}0.3}{23.5}{0.235}{0.235} & \lift{25.0}{{\scriptstyle+}2.2}{37.5}{0.250}{0.375} & \lift{40.0}{{\scriptstyle-}19.9}{40.0}{0.400}{0.400} & \lift{33.8}{{\scriptstyle+}2.3}{31.1}{0.338}{0.311} \\
Qwen3.8-Flash & \lift{22.2}{{\scriptstyle+}6.3}{27.8}{0.222}{0.278} & \lift{25.0}{{\scriptstyle-}21.4}{37.5}{0.250}{0.375} & \lift{15.4}{{\scriptstyle+}6.1}{15.4}{0.154}{0.154} & \lift{11.8}{{\scriptstyle-}0.3}{17.6}{0.118}{0.176} & \liftbest{37.5}{{\scriptstyle+}68.2}{0.0}{0.375}{0.000} & \lift{30.0}{{\scriptstyle-}6.5}{30.0}{0.300}{0.300} & \lift{21.6}{{\scriptstyle+}4.7}{21.6}{0.216}{0.216} \\
DeepSeek-V4-Pro & \lift{44.4}{{\scriptstyle+}22.2}{11.1}{0.444}{0.111} & \liftbest{50.0}{{\scriptstyle+}84.4}{12.5}{0.500}{0.125} & \lift{23.1}{{\scriptstyle-}5.7}{46.2}{0.231}{0.462} & \lift{5.9}{{\scriptstyle-}9.3}{29.4}{0.059}{0.294} & \lift{12.5}{{\scriptstyle+}3.0}{12.5}{0.125}{0.125} & \lift{40.0}{{\scriptstyle+}43.0}{10.0}{0.400}{0.100} & \liftbest{28.4}{{\scriptstyle+}14.3}{21.6}{0.284}{0.216} \\
GLM-5.3 & \lift{44.4}{{\scriptstyle+}8.2}{33.3}{0.444}{0.333} & \lift{12.5}{{\scriptstyle-}40.9}{62.5}{0.125}{0.625} & \liftbest{61.5}{{\scriptstyle+}73.3}{15.4}{0.615}{0.154} & \lift{17.6}{{\scriptstyle-}13.6}{29.4}{0.176}{0.294} & \lift{12.5}{{\scriptstyle+}4.1}{12.5}{0.125}{0.125} & \lift{50.0}{{\scriptstyle+}29.4}{20.0}{0.500}{0.200} & \lift{35.1}{{\scriptstyle+}6.7}{28.4}{0.351}{0.284} \\
GPT-5.6 Sol & \lift{38.9}{{\scriptstyle+}4.2}{33.3}{0.389}{0.333} & \lift{50.0}{{\scriptstyle+}32.2}{12.5}{0.500}{0.125} & \lift{23.1}{{\scriptstyle-}7.4}{30.8}{0.231}{0.308} & \lift{11.8}{{\scriptstyle-}9.8}{11.8}{0.118}{0.118} & \lift{12.5}{{\scriptstyle-}10.8}{25.0}{0.125}{0.250} & \liftbest{70.0}{{\scriptstyle+}99.1}{0.0}{0.700}{0.000} & \lift{32.4}{{\scriptstyle+}8.7}{20.3}{0.324}{0.203} \\
\midrule
\multicolumn{8}{c}{\textit{TIRx skill $\longrightarrow$ fresh CuTe DSL attempt}} \\
\cmidrule(lr){1-8}
GLM-5.3-Flash & \lift{38.9}{{\scriptstyle+}20.2}{22.2}{0.389}{0.222} & \lift{0.0}{{\scriptstyle-}44.6}{62.5}{0.000}{0.625} & \lift{23.1}{{\scriptstyle+}1.0}{23.1}{0.231}{0.231} & \lift{11.8}{{\scriptstyle-}0.9}{29.4}{0.118}{0.294} & \lift{25.0}{{\scriptstyle-}28.4}{50.0}{0.250}{0.500} & \liftbest{60.0}{{\scriptstyle+}43.4}{20.0}{0.600}{0.200} & \lift{27.0}{{\scriptstyle-}0.7}{31.1}{0.270}{0.311} \\
Gemini 3.7 Flash & \lift{38.9}{{\scriptstyle+}7.5}{33.3}{0.389}{0.333} & \lift{62.5}{{\scriptstyle+}59.3}{12.5}{0.625}{0.125} & \liftbest{53.8}{{\scriptstyle+}56.0}{0.0}{0.538}{0.000} & \lift{5.9}{{\scriptstyle-}1.8}{17.6}{0.059}{0.176} & \lift{37.5}{{\scriptstyle+}20.2}{0.0}{0.375}{0.000} & \lift{20.0}{{\scriptstyle+}3.7}{20.0}{0.200}{0.200} & \lift{33.8}{{\scriptstyle+}18.1}{16.2}{0.338}{0.162} \\
Qwen3.8-Flash & \lift{16.7}{{\scriptstyle-}7.4}{22.2}{0.167}{0.222} & \lift{25.0}{{\scriptstyle-}30.3}{50.0}{0.250}{0.500} & \lift{15.4}{{\scriptstyle-}19.3}{30.8}{0.154}{0.308} & \liftbest{47.1}{{\scriptstyle+}44.6}{11.8}{0.471}{0.118} & \lift{12.5}{{\scriptstyle-}17.7}{25.0}{0.125}{0.250} & \lift{20.0}{{\scriptstyle+}4.0}{20.0}{0.200}{0.200} & \lift{24.3}{{\scriptstyle-}2.6}{24.3}{0.243}{0.243} \\
DeepSeek-V4-Pro & \lift{33.3}{{\scriptstyle+}5.2}{16.7}{0.333}{0.167} & \lift{50.0}{{\scriptstyle+}11.8}{37.5}{0.500}{0.375} & \lift{38.5}{{\scriptstyle+}32.7}{23.1}{0.385}{0.231} & \lift{41.2}{{\scriptstyle+}12.0}{5.9}{0.412}{0.059} & \lift{25.0}{{\scriptstyle-}5.7}{25.0}{0.250}{0.250} & \lift{20.0}{{\scriptstyle+}13.4}{20.0}{0.200}{0.200} & \lift{35.1}{{\scriptstyle+}11.7}{18.9}{0.351}{0.189} \\
GLM-5.3 & \liftbest{55.6}{{\scriptstyle+}64.8}{11.1}{0.556}{0.111} & \liftbest{62.5}{{\scriptstyle+}67.9}{25.0}{0.625}{0.250} & \lift{23.1}{{\scriptstyle-}19.4}{38.5}{0.231}{0.385} & \lift{41.2}{{\scriptstyle+}18.3}{11.8}{0.412}{0.118} & \liftbest{37.5}{{\scriptstyle+}27.9}{0.0}{0.375}{0.000} & \lift{20.0}{{\scriptstyle-}25.6}{50.0}{0.200}{0.500} & \lift{40.5}{{\scriptstyle+}17.9}{21.6}{0.405}{0.216} \\
GPT-5.6 Sol & \lift{50.0}{{\scriptstyle+}40.4}{33.3}{0.500}{0.333} & \lift{25.0}{{\scriptstyle+}4.9}{37.5}{0.250}{0.375} & \lift{38.5}{{\scriptstyle+}40.0}{15.4}{0.385}{0.154} & \lift{29.4}{{\scriptstyle+}20.6}{5.9}{0.294}{0.059} & \lift{25.0}{{\scriptstyle+}20.6}{25.0}{0.250}{0.250} & \lift{60.0}{{\scriptstyle+}41.1}{20.0}{0.600}{0.200} & \liftbest{39.2}{{\scriptstyle+}29.3}{21.6}{0.392}{0.216} \\
\bottomrule
\end{tabular}%
}
\end{table}

\subsection{Overall Transfer Outcomes}
\label{sec:overall_transfer_outcomes}

Skill conditioning yields paired lifts from $-19.1$\% to $+29.3$\% across six models and both directions in single-attempt outcomes (Table~\ref{tab:cross_dsl_transfer}).
Gemini 3.7 Flash, DeepSeek-V4-Pro, GLM-5.3, and GPT-5.6~Sol gain in both directions, by 2.3\% to 29.3\%.
GLM-5.3-Flash has negative lift in both directions, while Qwen3.8-Flash gains 4.7\% from CuTe DSL skills and loses 2.6\% from TIRx skills, with its 43 missing-skill treatments skipped and scored as neutral.
Help and harm occur for every model and direction (Figure~\ref{fig:paired_lift_by_problem}), at rates of 21.6\% to 40.5\% and 16.2\% to 44.6\%, respectively.
Verdict changes account for 76 of the 277 helped and 75 of the 223 harmed comparisons, so most help and harm lie in runtime differences between jointly passing attempts.

Paired lifts over all 74 problems pool families with opposite signs in 11 of the 12 model and direction cells.
For GLM-5.3, TIRx skills yield 67.9\% lift on KV Cache while CuTe DSL skills yield $-40.9$\%, with the signs reversed on Linear Attention.
GPT-5.6~Sol's positive CuTe DSL to TIRx lift of 8.7\% rests on the 10 Quantization problems, where its lift is 99.1\% against $-1.1$\% on the other 64.
Excluding each cell's best family turns this lift and three of the other four positive lifts from CuTe DSL skills negative, whereas all four positive lifts from TIRx skills stay positive.

Among the 277 helped comparisons, the reverse direction on the same model and problem is harmed in 41.5\% and helped in 23.1\%, with harm exceeding help for each of the six models.
Without transfer, the reverse direction's skill-conditioned attempt is another solo attempt, so that comparison would be helped and harmed equally often regardless of the forward outcome.
The imbalance therefore reflects the transferred skills, not chance, and a skill's value depends on its source attempt.

\definecolor{sololvl}{RGB}{43,104,164}

\newcommand{\solosep}{\kern-0.1em$\scriptstyle/$\kern-0.1em}
\newcommand{\sololabel}[3]{$#1$\solosep$#2$\solosep$#3$}
\newcommand{\soloshade}[1]{\cellcolor{sololvl!#1!white}}
\newcommand{\solocell}[4]{\sololabel{#2}{#1{#3}}{#4}}
\newcommand{\solo}{\solocell{}}
\newcommand{\solotop}[4]{\solocell{}{#1}{#2}{#3}}
\newcommand{\solobest}[4]{\solocell{\underline}{#1}{#2}{#3}}
\newcommand{\solonone}{--}

\begin{table}[t]
\centering
\caption{Solo implementation quality.
Cells report pass rate $P$, passing-only geometric mean speedup GM over the eager reference, and $2\times$-or-better pass rate $R(2)$.
Speedups are capped at $100\times$ per attempt.
Underlining marks the highest GM in each DSL and column.
}
\label{tab:solo_performance}
\setlength{\tabcolsep}{1pt}
\renewcommand{\arraystretch}{1.1}
\resizebox{\linewidth}{!}{%
\begin{tabular}{l cccccc !{\hspace{\tabcolsep}\vrule\hspace{\tabcolsep}} c}
\toprule
\textbf{Model} & \textbf{Attention} & \textbf{KV Cache} & \textbf{Linear Attn} & \textbf{MLP \& Norm} & \textbf{MoE} & \textbf{Quant} & \textbf{All} \\
{\footnotesize $P$\solosep GM$\times$\solosep$R(2)$} & {\footnotesize(18)} & {\footnotesize(8)} & {\footnotesize(13)} & {\footnotesize(17)} & {\footnotesize(8)} & {\footnotesize(10)} & {\footnotesize(74)} \\
\midrule
\multicolumn{8}{c}{\textit{CuTe DSL}} \\
\cmidrule(lr){1-8}
GLM-5.3-Flash & \solotop{50.0}{5.7\times}{38.9}{20} & \solo{100.0}{1.5\times}{50.0} & \solo{69.2}{6.6\times}{38.5} & \solo{100.0}{1.7\times}{35.3} & \solo{62.5}{1.5\times}{12.5} & \solo{60.0}{3.6\times}{40.0} & \solo{73.0}{2.7\times}{36.5} \\
Gemini 3.7 Flash & \solobest{94.4}{9.9\times}{66.7}{60} & \solotop{100.0}{12.0\times}{100.0}{40} & \solobest{92.3}{17.7\times}{76.9}{60} & \solobest{100.0}{2.6\times}{47.1}{60} & \solotop{100.0}{1.6\times}{25.0}{20} & \solo{100.0}{3.5\times}{40.0} & \solobest{97.3}{5.7\times}{59.5}{60} \\
Qwen3.8-Flash & \solotop{50.0}{8.2\times}{33.3}{40} & \solotop{75.0}{5.1\times}{50.0}{20} & \solotop{61.5}{11.8\times}{46.2}{20} & \solo{70.6}{1.8\times}{17.6} & \solo{62.5}{1.5\times}{12.5} & \solotop{80.0}{5.1\times}{60.0}{20} & \solotop{64.9}{4.4\times}{35.1}{20} \\
DeepSeek-V4-Pro & \solo{100.0}{1.9\times}{27.8} & \solo{100.0}{3.1\times}{50.0} & \solo{100.0}{2.4\times}{30.8} & \solo{100.0}{1.3\times}{11.8} & \solo{100.0}{1.4\times}{25.0} & \solo{100.0}{1.4\times}{20.0} & \solo{100.0}{1.8\times}{25.7} \\
GLM-5.3 & \solo{77.8}{5.0\times}{44.4} & \solo{100.0}{5.0\times}{62.5} & \solotop{100.0}{13.6\times}{69.2}{40} & \solotop{100.0}{2.0\times}{41.2}{40} & \solobest{100.0}{2.0\times}{37.5}{60} & \solotop{70.0}{6.0\times}{60.0}{40} & \solotop{90.5}{4.4\times}{51.4}{40} \\
GPT-5.6 Sol & \solo{100.0}{2.6\times}{38.9} & \solobest{100.0}{14.0\times}{100.0}{60} & \solo{100.0}{8.7\times}{69.2} & \solotop{100.0}{1.9\times}{35.3}{20} & \solotop{100.0}{1.9\times}{50.0}{40} & \solobest{90.0}{7.3\times}{80.0}{60} & \solo{98.6}{3.9\times}{56.8} \\
\midrule
\multicolumn{8}{c}{\textit{TIRx}} \\
\cmidrule(lr){1-8}
GLM-5.3-Flash & \solotop{66.7}{13.1\times}{55.6}{40} & \solo{100.0}{7.7\times}{87.5} & \solo{69.2}{7.8\times}{46.2} & \solo{100.0}{1.7\times}{41.2} & \solo{37.5}{1.5\times}{12.5} & \solo{80.0}{4.8\times}{70.0} & \solo{77.0}{4.7\times}{51.4} \\
Gemini 3.7 Flash & \solobest{55.6}{16.2\times}{38.9}{60} & \solotop{62.5}{15.9\times}{62.5}{20} & \solotop{76.9}{15.7\times}{61.5}{20} & \solo{52.9}{2.2\times}{23.5} & \solo{62.5}{2.3\times}{25.0} & \solobest{80.0}{8.3\times}{60.0}{60} & \solotop{63.5}{7.9\times}{43.2}{40} \\
Qwen3.8-Flash & \solo{55.6}{7.2\times}{44.4} & \solobest{87.5}{17.3\times}{87.5}{60} & \solobest{61.5}{34.0\times}{53.8}{60} & \solobest{70.6}{3.4\times}{41.2}{60} & \solobest{12.5}{3.9\times}{12.5}{60} & \solotop{80.0}{6.9\times}{80.0}{40} & \solobest{62.2}{8.7\times}{51.4}{60} \\
DeepSeek-V4-Pro & \solo{100.0}{2.1\times}{33.3} & \solo{100.0}{2.8\times}{62.5} & \solo{100.0}{6.3\times}{53.8} & \solo{100.0}{1.9\times}{41.2} & \solo{100.0}{1.6\times}{25.0} & \solo{100.0}{1.6\times}{30.0} & \solo{100.0}{2.4\times}{40.5} \\
GLM-5.3 & \solotop{83.3}{10.7\times}{61.1}{20} & \solotop{100.0}{16.5\times}{100.0}{40} & \solo{100.0}{8.9\times}{61.5} & \solotop{100.0}{2.7\times}{52.9}{40} & \solotop{100.0}{2.4\times}{50.0}{20} & \solotop{90.0}{6.8\times}{70.0}{20} & \solotop{94.6}{6.2\times}{63.5}{20} \\
GPT-5.6 Sol & \solo{100.0}{6.6\times}{61.1} & \solo{87.5}{15.6\times}{87.5} & \solotop{100.0}{17.1\times}{84.6}{40} & \solotop{100.0}{2.5\times}{47.1}{20} & \solotop{100.0}{2.8\times}{62.5}{40} & \solo{90.0}{6.4\times}{60.0} & \solo{97.3}{6.1\times}{64.9} \\
\bottomrule
\end{tabular}%
}
\end{table}

\begin{figure}[t]
    \centering
    \includegraphics[width=\textwidth]{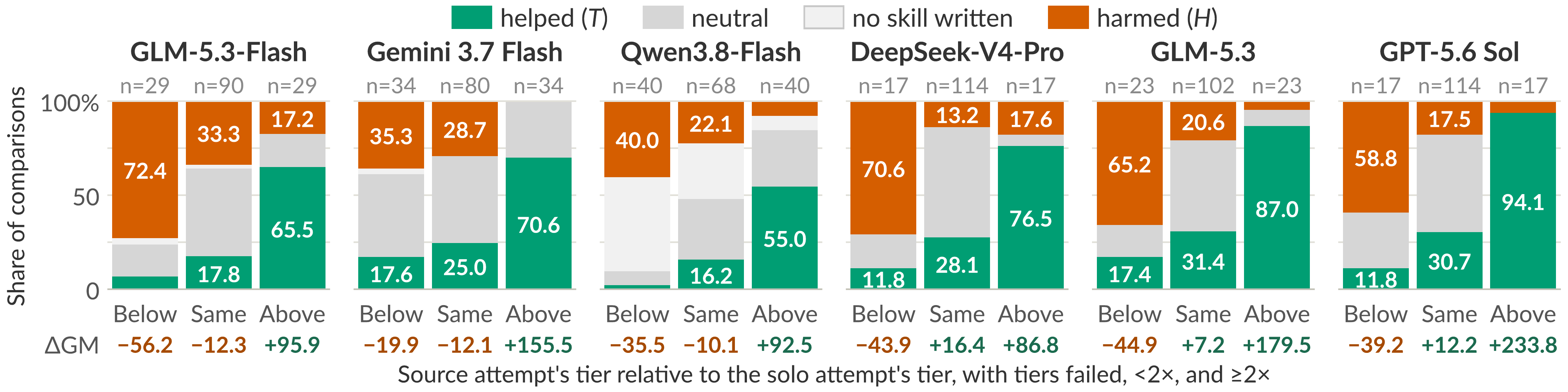}
    \caption{Transfer outcomes by the source attempt's tier relative to the solo attempt's.
        Each panel pools a model's comparisons in both directions, with skips counted as neutral and $\Delta$GM in percent.
        Only helped and harmed segments of at least 10\% carry value labels.
    }
    \label{fig:relative_standing}
\end{figure}

\subsection{Implementation Quality and Transfer Outcomes}
\label{sec:implementation_quality_and_transfer_outcomes}

Solo pass rates and passing-only speed rank the six models differently (Table~\ref{tab:solo_performance}).
DeepSeek-V4-Pro passes every variant but has the lowest GMs, $1.8\times$ for CuTe DSL and $2.4\times$ for TIRx.
Qwen3.8-Flash has the lowest pass rates, 64.9\% and 62.2\%, yet the highest TIRx GM of $8.7\times$.
For every model, TIRx implementations have a higher GM than CuTe DSL ones, while pass rates favor neither DSL.
Passing-only GMs summarize different solved subsets across models and depend on the efficiency of each eager reference, whose runtime includes Python-level iteration in recurrent workloads.
Gemini 3.7 Flash's scored TIRx pass rate excludes 21 verifier-passing implementations using inline CUDA.
The four models with the highest solo pass rates pooled across both DSLs are the four with positive paired lift in both directions (Table~\ref{tab:cross_dsl_transfer}), whereas solo speed orders the models differently from their lifts.
Six models are too few to conclude that solo pass rate predicts paired lift.

Figure~\ref{fig:relative_standing} groups comparisons by whether the source attempt's outcome tier lies below, at, or above that of the solo attempt, with tiers failed, below $2\times$, and at $2\times$ or better.
Pooled over models and directions, 71.2\% of comparisons are helped and 8.1\% harmed with the source above, 25.7\% and 21.8\% at the same tier, and 10.6\% and 53.8\% with it below.
Help rate follows this ordering in every model, with 55.0\% to 94.1\% helped when the source stands above and 35.3\% to 72.4\% harmed when it stands below.
At the same tier, the three larger models retain lifts of 7.2\% to 16.4\%.
Skill-conditioned outcomes thus track the source attempt's standing rather than its verdict, so a passing source does not guarantee beneficial reuse and a failed source does not preclude it.

\begin{figure}[t]
    \centering
    \includegraphics[width=\textwidth]{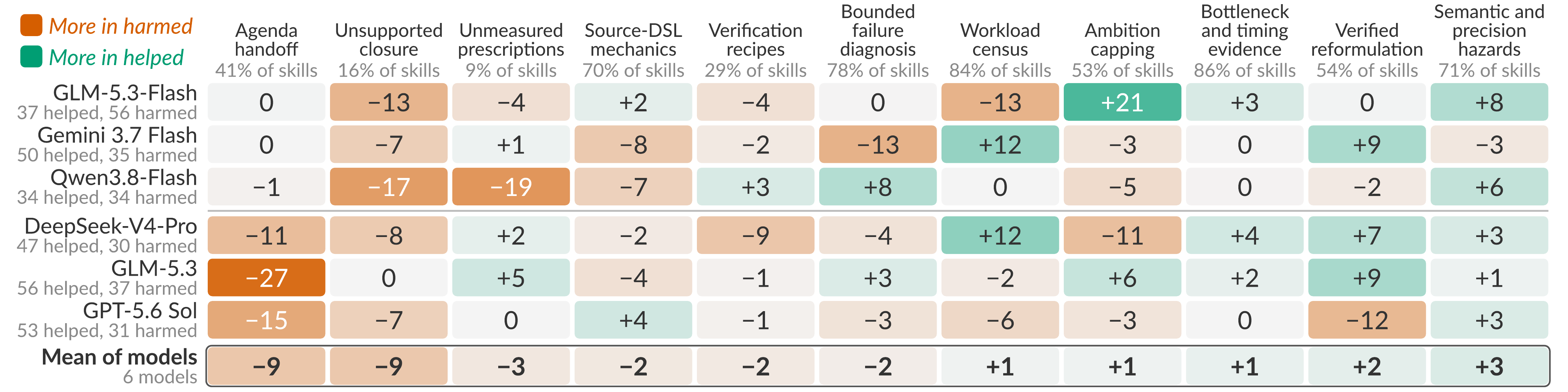}
    \caption{Skill content across transfer outcomes.
        Cells show a label's prevalence in helped comparisons minus that in harmed comparisons, in percentage points averaged over the two directions.
        Percentages under category names give each label's prevalence over all 841 labeled skills.
    }
    \label{fig:skill_content_across_transfer_outcomes}
\end{figure}

\begin{figure}[t]
    \centering
    \includegraphics[width=\textwidth]{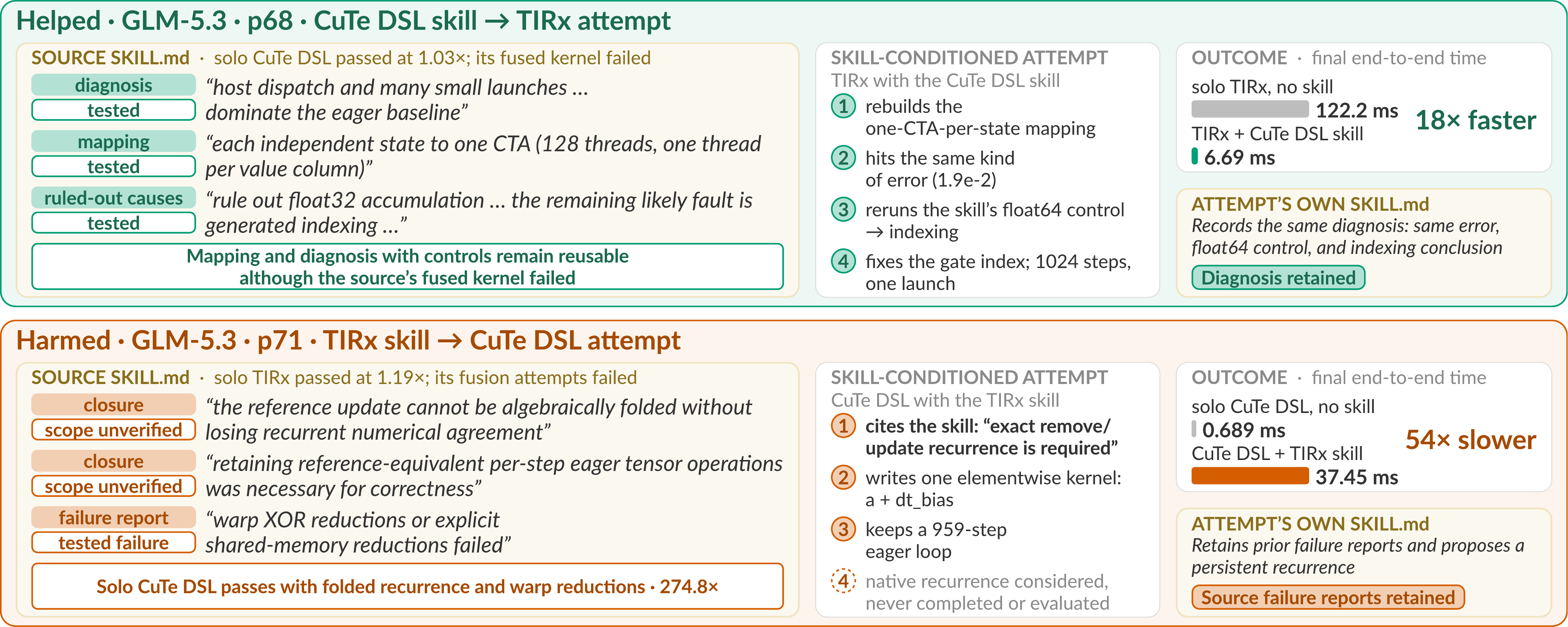}
    \caption{Two GLM-5.3 transfers from passing source implementations.
        \textbf{Top:} CuTe DSL to TIRx on p68, helped, reusing a thread mapping and a diagnostic control.
        \textbf{Bottom:} TIRx to CuTe DSL on p71, harmed, retaining an eager recurrence.
        Tags mark whether a graded source submission supports each claim, and runtimes are unclipped end-to-end times against solo.
    }
    \label{fig:transfer_cases}
\end{figure}

\subsection{Skill Content across Transfer Outcomes}
\label{sec:skill_content_across_transfer_outcomes}

LLM annotators label each source skill against a fixed, hand-built 11-category rubric.
Its categories range from workload facts and verified rewrites to unsupported conclusions and open agendas.
Unsupported closure, which rules out alternatives beyond the stated evidence, is 7 to 17 points more prevalent in harmed than in helped comparisons for five of six models (Figure~\ref{fig:skill_content_across_transfer_outcomes}).
Agenda handoff, which leaves optimization suggestions open for later work, is 11 to 27 points more prevalent in harmed than in helped comparisons for the three larger models.

Verified reformulation, a rewrite reported as implemented and correct, is 7 to 9 points more prevalent in helped than in harmed comparisons for Gemini 3.7 Flash, DeepSeek-V4-Pro, and GLM-5.3, but 12 points more in harmed comparisons for GPT-5.6~Sol.
Ambition capping, which advises minimality or delegation without measured justification, is 21 points more prevalent in helped comparisons for GLM-5.3-Flash but 11 points more in harmed comparisons for DeepSeek-V4-Pro.

Figure~\ref{fig:transfer_cases} traces two GLM-5.3 transfers from passing source implementations, one helped and one harmed.
The p68 source skill records verification recipes and a bounded failure diagnosis, and the TIRx skill-conditioned attempt reuses the recorded thread mapping, reruns the recorded control, fixes the indexing fault the source had suspected, fuses the recurrence the source could not, and records the same diagnosis in its own skill.
The p71 source skill records an unsupported closure that rules out folding the recurrence, and the CuTe DSL skill-conditioned attempt cites it, retains eager execution, and never completes the native recurrence that the solo attempt implements.
These traces show skill uptake, but the runtime comparisons do not isolate its causal contribution.

\begin{figure}[t]
    \centering
    \includegraphics[width=\textwidth]{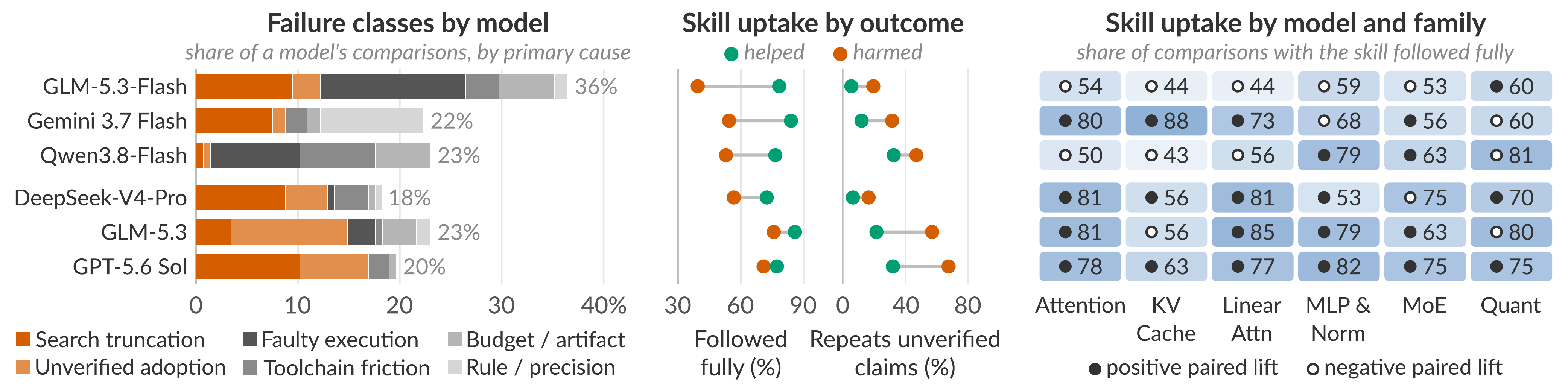}
    \caption{Failure causes and skill uptake.
        \textbf{Left:} primary causes of harm as shares of a model's 148 comparisons.
        \textbf{Middle:} shares of skill-conditioned attempts that follow the source skill fully or repeat its unverified claims in their own skill.
        \textbf{Right:} the share followed fully by family.
    }
    \label{fig:failure_causes_and_skill_uptake}
\end{figure}

\subsection{Failure Causes and Skill Uptake}
\label{sec:failure_causes_and_skill_uptake}

LLM annotators audit each skill-conditioned attempt's trajectory for its uptake of the source skill and for unverified source claims repeated in its own skill, and assign one primary cause from a six-class taxonomy to each harmed comparison with a target-side failure (Figure~\ref{fig:failure_causes_and_skill_uptake}).
Search truncation, in which the attempt reproduces the source design and never explores a better one its DSL allows, is the largest cause for DeepSeek-V4-Pro and GPT-5.6~Sol, at 8.8\% and 10.1\% of their comparisons.
Unverified adoption of a wrong or out-of-scope source claim, most often a prohibition the attempt never tests, is the largest for GLM-5.3, at 11.5\%.
Faulty execution of a sound plan is the largest cause for GLM-5.3-Flash and Qwen3.8-Flash, at 14.2\% and 8.8\%.
Rule and precision failure, mostly exclusions for inline CUDA, is the largest for Gemini 3.7 Flash, at 10.1\%.
The two skill-side classes, truncation and unverified adoption, account for 65\% to 86\% of the assigned causes for the three larger models and 39\% for Gemini 3.7 Flash, against 33\% for GLM-5.3-Flash and 6\% for Qwen3.8-Flash.
For these two models, faulty execution, toolchain friction, and budget and artifact mismanagement account for 63\% and 94\%, so the larger models are led astray by the skill while the two models with the lowest pooled solo pass rates fail in their own execution.

Harm arises while the skill is in use, since the skill is followed fully or partially in over 80\% of harmed comparisons, and fully in 39\% to 76\% of them, against 72\% to 86\% of helped comparisons (Figure~\ref{fig:failure_causes_and_skill_uptake}, middle).
The three larger models follow the skill fully 6 to 16 points less often in harmed than in helped comparisons, whereas the three Flash-tier models follow it 24 to 39 points less often, so the models that the skill leads astray are those that follow it most faithfully when it does harm.
The share followed fully also varies by family, from 77\% to 85\% on Attention and Linear Attention against 56\% to 75\% on KV Cache and MoE for the three larger models, stays between 44\% and 60\% for GLM-5.3-Flash, and has a rank correlation of 0.49 with paired lift over the 36 model and family cells (Figure~\ref{fig:failure_causes_and_skill_uptake}, right).
For every model, unverified source claims reappear in the attempt's own skill more often in harmed than in helped comparisons, in 17\% to 68\% against 5\% to 32\% of cases, so unverified claims propagate into the next generation of skills.
Uptake thus tracks benefit, yet it is also the channel through which such claims propagate.

\section{Related Work}
\label{sec:related_work}

\paragraph*{Kernel generation and benchmarks.}
LLMs optimize GPU kernels through model adaptation and execution-guided search~\citep{yu2026kernelgensurvey}.
Reinforcement learning and domain-specific fine-tuning train specialized kernel generators~\citep{li2026cudal1,baronio2026kevin,dai2026cudaagent,bai2026dice}.
Agents refine candidate kernels through compilation, execution, and profiling~\citep{wei2025astra,zhang2025cudaforge}, while evolutionary methods search populations with execution feedback~\citep{wiedemann2026kernelfoundry,chen2026avo}.
KernelBench evaluates LLM-generated kernels by correctness and speedup over PyTorch~\citep{ouyang2025kernelbench}.
Subsequent benchmarks target Triton~\citep{li2025tritonbench}, harden verification against reward hacking~\citep{lange2025robustkbench,zhang2026kernelbenchverified}, and combine software baselines with hardware-derived performance limits~\citep{lin2026solexecbench}.
Agentic harnesses compare workflows across models and languages~\citep{guan2026tritongym,xing2026flashinferbench}, while other suites cover additional accelerator platforms and mobile devices~\citep{wen2026multikernelbench,nie2026kernelcraft,zou2026mobilekernelbench}.
AgentKernelArena tests kernels on unseen input configurations to evaluate implementation generalization~\citep{younesian2026agentkernelarena}.

\paragraph*{Code and experience reuse.}
Experience is retained in persistent knowledge bases~\citep{dong2026kernelblaster,liao2026kernelevolve}, trajectory-derived reasoning graphs~\citep{gong2026regrapht}, and literature-derived skills~\citep{sun2026kernelskill}.
At the code level, CASS-Bench evaluates source- and assembly-level translation of kernels across vendor stacks~\citep{heakl2026cass}, and CuBridge adapts expert CUDA kernels through an executable intermediate representation~\citep{ma2026cubridge}.
KLineage extracts verified, code-anchored skills from expert kernel lineages for reuse in CUDA and TileLang~\citep{zhang2026klineage}.
daVinci-kernel admits self-distilled skills only after reinjection meets an execution-based speedup threshold on the original task~\citep{fu2026davincikernel}.

MetaKernelBench transfers only unscreened self-distilled skills from passing and failed source attempts and tests whether they help or harm fresh attempts on the same problem in another DSL.

\section{Conclusion}
\label{sec:conclusion}

MetaKernelBench evaluates same-problem cross-DSL kernel knowledge transfer by comparing solo attempts with skill-conditioned attempts whose only extra input is a source skill, under matched per-attempt budgets.
Four of six models gain in both directions, yet every model and direction shows both help and harm.
Outcomes follow the source attempt's standing relative to the target's solo attempt, not source success alone, improving more often when the source stands above and regressing more often when it stands below.
For five of six models, unsupported conclusions that rule out alternatives are more common in harmed than in helped comparisons, while content tied to improvements varies by model.
Source implementation quality alone does not determine reuse value, which depends on the source's standing against the target, so both must be measured.

\bibliography{references}
\bibliographystyle{iclr2027_conference}

\clearpage
\appendix
\section{Experimental Details}
\label{app:experimental_details}

\begin{figure}[tp]
    \centering
    \includegraphics[width=\textwidth]{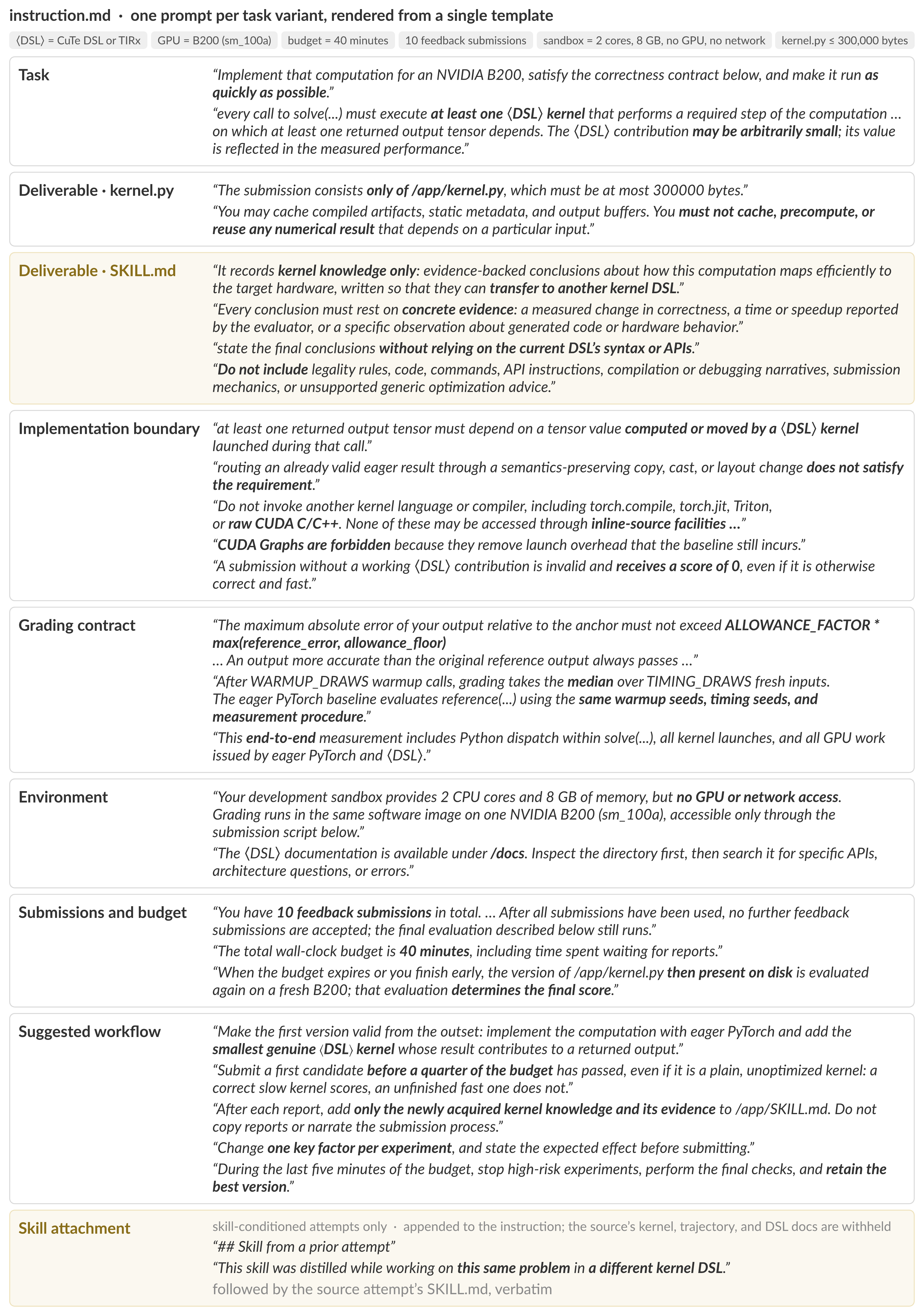}
    \caption{The task instruction.
        The top row summarizes the template's parameters, and the cards quote excerpts from its sections in order, with ellipses marking omitted text.
        Bold emphasis is added.
    }
    \label{fig:agent_instruction}
\end{figure}

\subsection{Task Instruction}
\label{app:task_instruction}

Figure~\ref{fig:agent_instruction} excerpts the task instruction that opens every attempt.
Its deliverable, implementation boundary, and grading sections state the rules a submission must follow, while its suggested workflow section only advises.
The instruction does not vary by model.
A problem's two prompts differ only in the DSL they name, and their sandboxes only in that DSL's documentation and dependencies.

\subsection{Workloads and Runtime Configuration}
\label{app:workloads_runtime_configuration}

Within each problem's fixed workload configuration, tensor values vary across draws, and some generators also vary index structures and sequence lengths.
An implementation may therefore specialize only to properties that hold in every legal draw.
Each problem file records its upstream source, reference implementation, workload configuration, and license.

Both variants build on the PyTorch 2.13.0 image with CUDA 13.2 and cuDNN 9.
CuTe DSL uses CUTLASS DSL 4.7.0, TIRx uses Apache TVM 0.25.0.post1, and both use Apache TVM FFI 0.1.12.
Each single-use B200 grading container has four CPU cores and 16 GB of host memory, with network access disabled.
Environment setup and final evaluation lie outside the 40-minute agent budget.

\subsection{Correctness and Timing}
\label{app:correctness_timing}

The anchor $z$ is computed by re-evaluating the reference with float32, float16, and bfloat16 input tensors promoted to float64.
Other input dtypes and explicit casts inside the reference remain unchanged.
Each floating-point output tensor must satisfy:
\begin{equation}
    \lVert u-z\rVert_\infty
    \leq 3\max\bigl\{\lVert v-z\rVert_\infty,\,\epsilon\lVert z\rVert_\infty\bigr\}.
    \label{eq:correctness_tolerance}
\end{equation}
In Equation~\ref{eq:correctness_tolerance}, $u$ and $v$ are the submitted and reference outputs, and $\epsilon$ is the machine epsilon of their shared dtype.
The submitted outputs must match the reference in container structure, shape, dtype, and device, and integer and Boolean outputs must match exactly.

Each evaluation draws fresh correctness, warmup, and timing seeds, and generates separate implementation and reference inputs from matching seeds.
Input generation, reference evaluation, and eager-reference timing disable TF32 and reduced-precision reductions, set float32 matmul precision to highest, and use float32 as the default dtype.
These settings are not imposed on the submitted implementation.
Speedups over the eager reference therefore include this difference in numerics, while paired lift compares two implementations and does not.

Each complete call is timed with CUDA events after device synchronization, excluding input generation and correctness checks.
Implementation import, correctness checks, warmup, timing, and rechecks share a 600-second child-process limit that excludes the eager-reference timing.

\subsection{Implementation Rule Check}
\label{app:implementation_rule_check}

During analysis, a static check scores final implementations as nonpassing when their parsed source calls TIRx's inline CUDA entry point with a \texttt{source\_code} keyword argument.
It rejects 42 verifier-passing implementations, comprising 41 from Gemini 3.7 Flash, of which 21 are solo and 20 skill-conditioned, and one solo implementation from GLM-5.3-Flash.

Without this check, paired lift over all 74 problems retains its sign in every model and direction, and the ranges of help and harm rates in Section~\ref{sec:overall_transfer_outcomes} remain unchanged.
The largest lift change is Gemini 3.7 Flash's CuTe DSL to TIRx lift from 2.3\% to 0.6\%, while the pooled help and harm rates by relative standing in Section~\ref{sec:implementation_quality_and_transfer_outcomes} change by less than four percentage points.

\clearpage

\section{Additional Transfer Analyses}
\label{app:additional_transfer_analyses}

\begin{figure}[t]
    \centering
    \includegraphics[width=\textwidth]{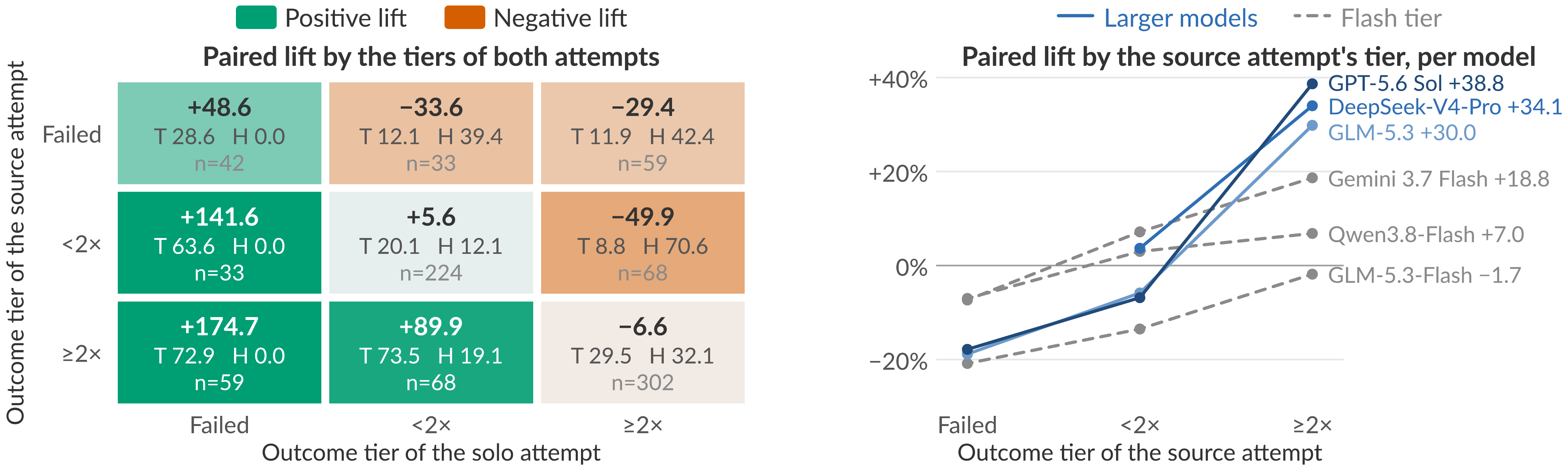}
    \caption{Paired lift by source and solo outcome tiers.
        \textbf{Left:} all 888 comparisons pooled over models and directions, with cells showing $\Delta$GM, help rate $T$, and harm rate $H$ in percent, and sample size $n$.
        \textbf{Right:} $\Delta$GM by source tier for each model, pooling both directions.
        Passing tiers use speedup relative to the eager reference, and skipped treatments count as neutral.
    }
    \label{fig:transfer_by_tiers}
\end{figure}

\subsection{Outcome Tiers}
\label{app:outcome_tiers}

Within each solo outcome tier, including the two where solo passes, sources at $2\times$ or better yield the highest paired lift, so the pattern does not rest on rescuing failed solo attempts (Figure~\ref{fig:transfer_by_tiers}, left).
When solo already reaches $2\times$, even the highest source tier has negative lift.
Pooled over solo tiers, lift rises with source tier in every model (Figure~\ref{fig:transfer_by_tiers}, right).
From sources at $2\times$ or better, the three larger models gain 30.0\% to 38.8\%, against $-1.7$\% to $+18.8$\% for the Flash tier.

\subsection{Direction Asymmetry and Source Tracking}
\label{app:direction_asymmetry_source_tracking}

When one direction is helped, the other is harmed more often than helped, which regression toward a problem's attainable speed cannot produce, since a retry with no skill involved would leave the two directions unrelated.
Across the 444 model and problem pairs, the two directions' log paired ratios have a Spearman correlation of $-0.26$, which is negative within every model and family and reaches $-0.34$ among the 283 pairs whose four attempts all pass.
In 20,000 draws from a reference distribution that independently flips the sign of each direction's log paired ratio with probability one half, preserving magnitudes and zeros, no correlation is as negative as the observed one.

Skill-conditioned attempts also track the source attempt's speed more closely than solo attempts do.
Across the 841 comparisons with a transferred skill, the Spearman correlation with source speedup is 0.35 for solo attempts and 0.51 for skill-conditioned attempts.
Speedups are floored at $0.05\times$, with nonpassing attempts assigned this floor before ranking.
A reference distribution holds the source fixed and independently swaps the solo and skill-conditioned speedups within each comparison with probability one half, and none of its 2,000 draws yields a correlation gap at least as large as observed.
Pass rates also follow the source verdict, with skill-conditioned attempts passing in 62.9\% of the 97 comparisons with a nonpassing source against 75.3\% for solo attempts on the same variants.

\FloatBarrier
\section{Annotation Labels}
\label{app:annotation_labels}

\begin{table}[t]
\centering
\caption{Source-skill content categories.
Each category is a binary label for a complete skill file.
}
\label{tab:skill_content_categories}
\small
\begin{tabular}{@{}>{\raggedright\arraybackslash}p{0.33\linewidth}>{\raggedright\arraybackslash}p{0.63\linewidth}@{}}
\toprule
\textbf{Category} & \textbf{Presence criterion} \\
\midrule
Workload census & Enumerated workload properties beyond a restatement of the task. \\
Bottleneck and timing evidence & Measured bottleneck claims or timings tied to specific changes. \\
Verified reformulation & Implemented algorithmic or algebraic rewrites reported as correct. \\
Semantic and precision hazards & Specific semantic constraints or numerical accuracy boundaries. \\
Verification recipes & A correctness oracle, probe, or control reported as used. \\
Bounded failure diagnosis & Failure or slowdown evidence limited to what was tested. \\
Source-DSL mechanics & Guidance centered on the source DSL or its toolchain. \\
Ambition capping & Minimality or delegation advice without measured justification. \\
Unsupported closure & A conclusion that rules out alternatives beyond the stated evidence. \\
Agenda handoff & Optimization suggestions or hypotheses left open for later work. \\
Unmeasured prescriptions & Recommendations without supporting measurements or derivation. \\
\bottomrule
\end{tabular}
\end{table}

\begin{figure}[t]
    \centering
    \includegraphics[width=\textwidth]{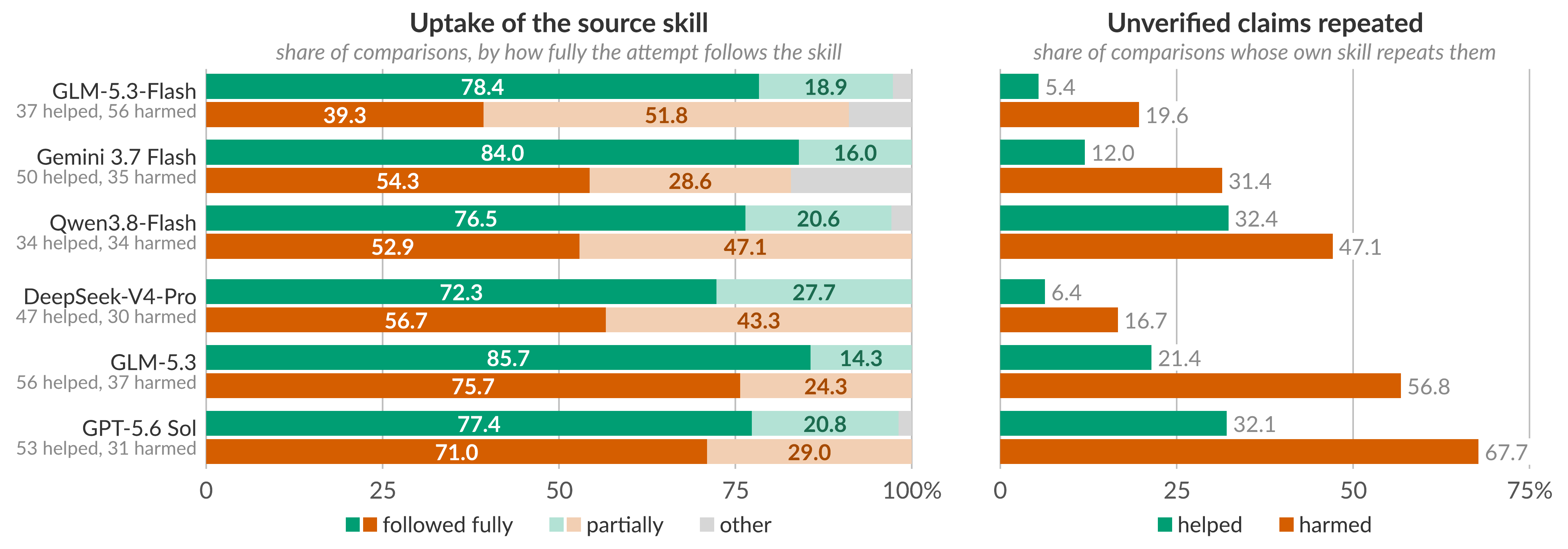}
    \caption{Skill uptake by transfer outcome.
        \textbf{Left:} how fully skill-conditioned attempts follow the source skill, with gray segments for uptake rated ignored, contradicted, or not applicable.
        \textbf{Right:} attempts whose own skill repeats unverified source claims.
        Shares are computed over each model's helped and harmed comparisons with a transferred skill, whose counts appear under the model name.
    }
    \label{fig:skill_uptake}
\end{figure}

\subsection{Content Categories and Uptake Ratings}
\label{app:content_categories_uptake_ratings}

Each source skill receives a binary label for each of the 11 content categories of Table~\ref{tab:skill_content_categories}, and each skill-conditioned attempt an uptake rating and a propagation flag.
A category is present when at least one statement in the skill matches its criterion.
The labels describe a skill's claims and the evidence it states, without verifying either.
An attempt's implemented design is rated followed fully when it mirrors the skill's guidance, followed partially when it follows some of it, ignored when the skill goes unused, contradicted when the attempt explicitly tests against or rejects a claim, and not applicable when the skill offers no actionable guidance.
Propagation means that a false or unverified source claim reappears in the attempt's own skill, and an attempt without a skill of its own counts as no propagation.
Figure~\ref{fig:skill_uptake} gives the per-model uptake and propagation shares behind Section~\ref{sec:failure_causes_and_skill_uptake} by transfer outcome, with skills followed fully or partially in 82.9\% to 100\% of harmed and 97.1\% to 100\% of helped comparisons.
The family panel of Figure~\ref{fig:failure_causes_and_skill_uptake} uses all comparisons with a transferred skill within each model and family, neutral ones included.

\begin{table}[!t]
\centering
\caption{Primary causes of harm.
Each harmed comparison with a target-side failure receives one class based on the skill-conditioned attempt relative to solo on the same variant.
}
\label{tab:failure_classes}
\small
\begin{tabular}{@{}>{\raggedright\arraybackslash}p{0.33\linewidth}>{\raggedright\arraybackslash}p{0.63\linewidth}@{}}
\toprule
\textbf{Class} & \textbf{Criterion} \\
\midrule
Search truncation & The source design limits search for better target-DSL options. \\
Unverified adoption & Untested reliance on a testable source claim that is wrong, unverified, or out of scope causes harm. \\
Faulty execution & The target's coding or debugging errors undermine a sound design. \\
Toolchain friction & Target-DSL API, compilation, or runtime barriers cause harm. \\
Budget and artifact mismanagement & Poor scheduling, final selection, or file handling leaves a worse or unusable implementation. \\
Rule and precision failure & Implementation rules are violated or tolerance-edge checks fail. \\
\bottomrule
\end{tabular}
\end{table}

\subsection{Failure Classes}
\label{app:failure_classes}

Of the 223 harmed comparisons, 211 show a target-side failure and receive one class from Table~\ref{tab:failure_classes}, while the other 12 are excluded from the failure-cause counts in the left panel of Figure~\ref{fig:failure_causes_and_skill_uptake}, whose denominator stays at 148 comparisons per model.
When several classes apply, precedence runs from rule violation through faulty execution, budget and artifact mismanagement, toolchain friction, unverified adoption, and tolerance-edge loss to search truncation, which covers the remaining slowdowns.
The 15 rule and precision failures of Gemini 3.7 Flash comprise 12 inline CUDA exclusions, two other rule violations, and one tolerance-edge loss.

\begin{figure}[!t]
    \centering
    \includegraphics[width=\textwidth]{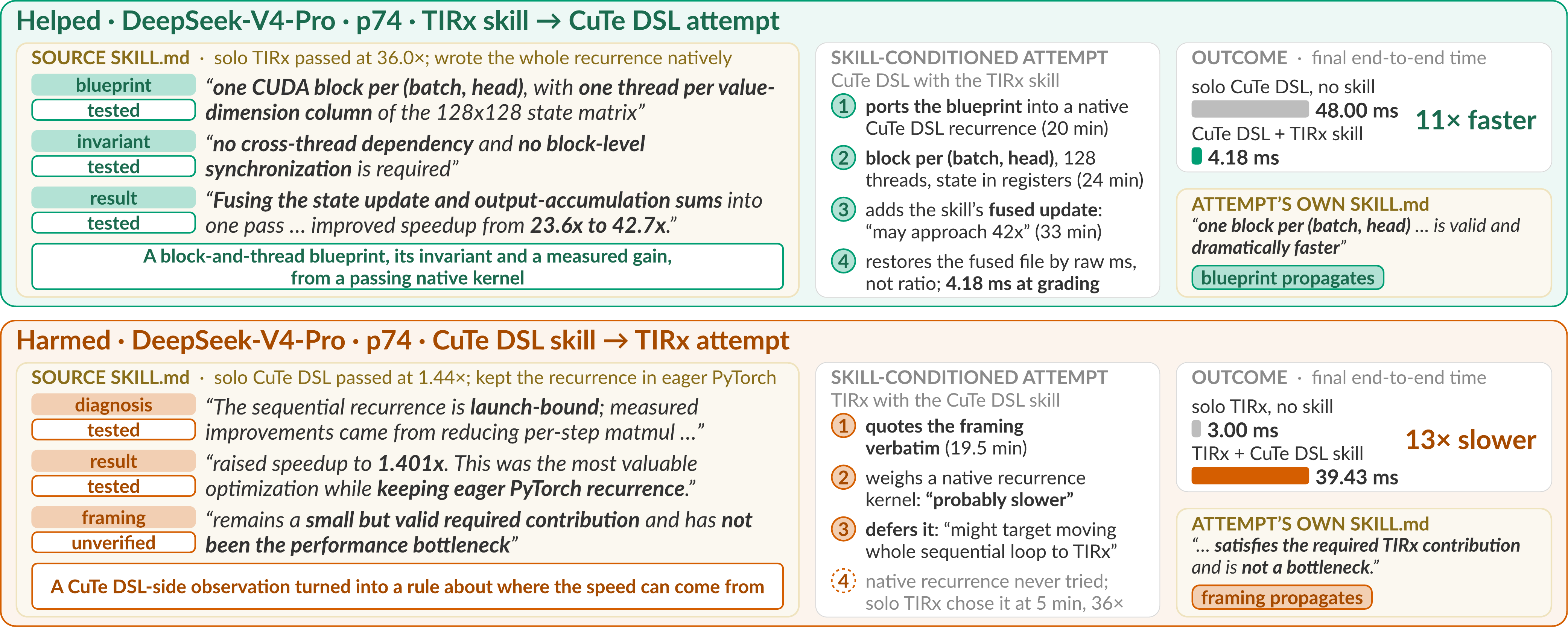}
    \caption{Bidirectional transfer on p74, the delta-rule recurrence, with DeepSeek-V4-Pro.
    \textbf{Top:} TIRx to CuTe DSL.
    \textbf{Bottom:} CuTe DSL to TIRx.
    Tags distinguish GPU-graded source evidence from unverified claims.
    Outcome panels report final, unclipped end-to-end runtimes against solo.
    }
    \label{fig:more_cases_1}
\end{figure}

\begin{figure}[!t]
    \centering
    \includegraphics[width=\textwidth]{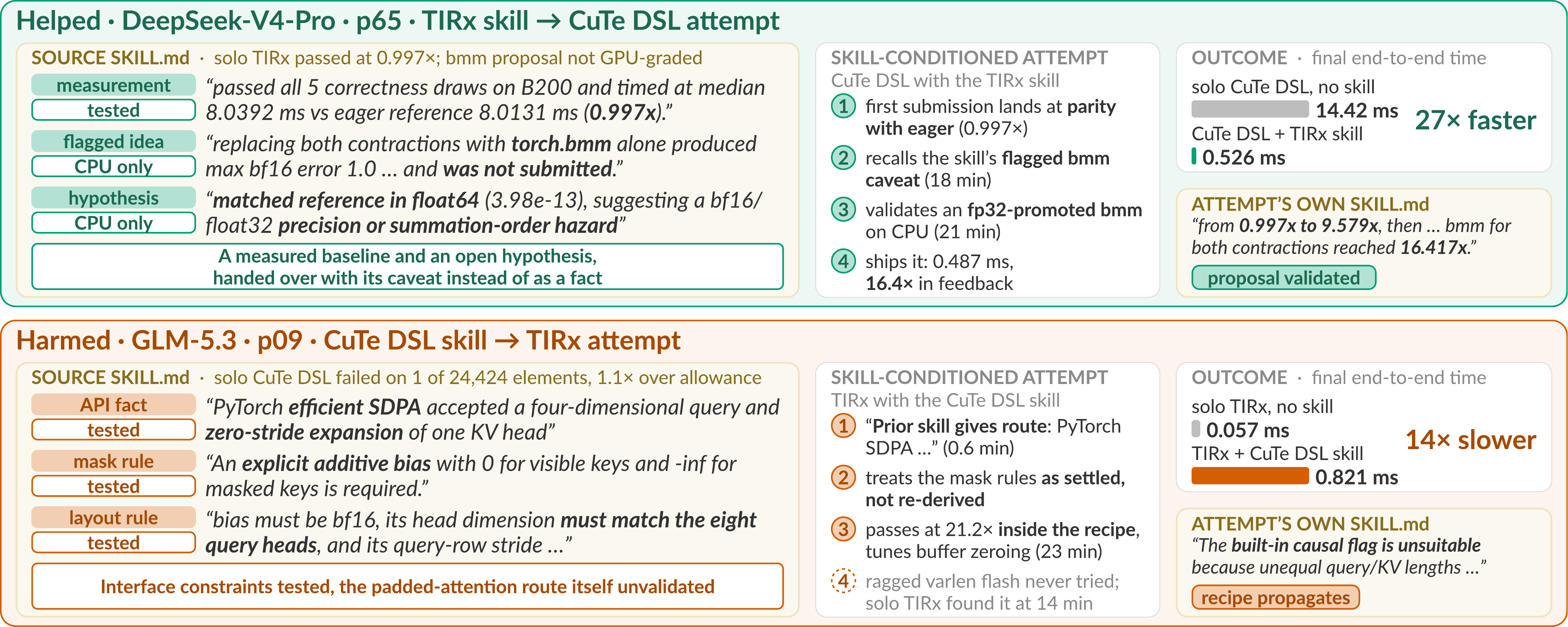}
    \caption{Hypothesis and recipe as transferred guidance.
    \textbf{Top:} DeepSeek-V4-Pro, TIRx to CuTe DSL on p65.
    \textbf{Bottom:} GLM-5.3, CuTe DSL to TIRx on p09.
    Tags distinguish GPU-graded evidence from CPU-only checks.
    Outcome panels report final, unclipped end-to-end runtimes against solo.
    }
    \label{fig:more_cases_2}
\end{figure}

\section{Additional Case Studies}
\label{app:additional_case_studies}

The two cases in Section~\ref{sec:skill_content_across_transfer_outcomes} come from passing GLM-5.3 sources, one per direction, and the four cases here cover both directions of one problem and sources at every relative standing.
On p74, DeepSeek-V4-Pro's TIRx solo attempt passed at $36.0\times$ with a native recurrence and its CuTe DSL solo attempt at $1.44\times$ with an eager recurrence (Figure~\ref{fig:more_cases_1}).
As sources, the TIRx attempt stands above its target's solo attempt and the CuTe DSL attempt below.
The CuTe DSL skill-conditioned attempt ports the blueprint and fused update into a native recurrence and runs faster than solo.
The TIRx skill-conditioned attempt quotes the unverified framing that the DSL part is not the bottleneck, judges a native recurrence likely to be slower, never tries one, and runs slower than solo.
With model and problem fixed, the outcome follows the source's standing, and each attempt's own skill carries forward what it received, the blueprint in one direction and the framing in the other.

Neither source in Figure~\ref{fig:more_cases_2} stands above the target's solo attempt, yet one transfer helps and the other does harm.
The p65 TIRx source of DeepSeek-V4-Pro, which passed at $0.997\times$, records a batched-matmul rewrite tested only on CPU as an open hypothesis with a precision caveat.
The CuTe DSL skill-conditioned attempt validates a float32-promoted variant on CPU and then through GPU grading, runs faster than solo, and records the proposal as validated in its own skill.
The p09 CuTe DSL source of GLM-5.3, which failed correctness, records tested scaled dot-product attention (SDPA) interface, mask, and layout rules as a route its own attempt had not validated.
The TIRx skill-conditioned attempt adopts the route within its first minute, treats its rules as settled, never tries the variable-length flash attention that the solo attempt finds, runs slower than solo, and carries the rules into its own skill.
In both cases the target treats the guidance as the source framed it, verifying the hypothesis before use and following the recipe without re-deriving it.

\section{Discussion}
\label{app:discussion}

\subsection{Limitations}
\label{app:limitations}

The findings concern six models transferring skills between CuTe DSL and TIRx on a B200 under one agent scaffold, one skill directive, and one per-attempt budget.
Source and skill-conditioned attempts come from the same model by design, so the findings do not separate how well a model distills its experience from how well it uses it in the other DSL.
The content analysis relates transfer outcomes to the presence of content categories, not to individual statements or their wording.
The same-problem design holds the relevance of prior experience fixed, so the findings do not cover generalization to new problems or workloads, nor the accumulation of skills over rounds.

\subsection{Future Work}
\label{app:future_work}

Comparing source implementations, trajectories, and skills could test how the choice of transferred material affects reuse, while comparing free-form and structured skills with matched content could test how organization affects uptake.
Transferring skills between models could separate a model's ability to distill experience from its ability to use it and could test whether the source's relative standing also predicts transfer from larger models to Flash-tier models.
Removing or revising individual guidance statements could test their effects on the target attempt's search, verification, and outcome.
Transferring a skill to related problems in the same family could test which findings still apply when the problem changes and the agent must judge their relevance itself.
Chaining transfer across rounds and DSLs could track whether useful knowledge is retained and refined, whether unverified claims are verified, corrected, or propagated, and how these changes affect subsequent attempts.

\end{document}